%% file: main.tex
\PassOptionsToPackage{table}{xcolor}
\documentclass[letterpaper]{article} 
\usepackage[preprint]{aaai2027}
\usepackage[hyphens]{url} 
\usepackage{graphicx} 
\usepackage[percent]{overpic}
\usepackage{natbib} 
\usepackage{caption} 
\usepackage{amsmath,amssymb,booktabs,multirow,tabularx,xspace}
\usepackage{placeins}
\usepackage{multicol}
\usepackage{xcolor}
\definecolor{TableHeader}{RGB}{229,230,239}
\definecolor{TableStripe}{RGB}{247,247,249}
\definecolor{PaceHighlight}{RGB}{255,247,205}
\definecolor{FigureBlue}{RGB}{18,75,174}
\newcolumntype{Y}{>{\centering\arraybackslash}X}
\newcommand{\method}{PACE\xspace}

\title{PACE: Propagation-Aware Collaborative Correction for One-Shot Personalized Federated Graph Learning}
\author{Ruizhe Huang\equalcontrib, Chengran Li\equalcontrib, Xiaochuan Shi\corresponding}
\affiliations{
School of Cyber Science and Engineering, Wuhan University\\
\texttt{ruizhehuang@whu.edu.cn, lichengran0@whu.edu.cn, shixiaochuan@whu.edu.cn}
}

\begin{document}
\maketitle
\begin{abstract}
\input{sec/abstract}
\end{abstract}
\input{sec/1_introduction}
\input{sec/2_related_work}
\input{sec/3_problem}
\input{sec/4_method}
\input{sec/5_experiments}
\input{sec/6_conclusion}
\begin{small}
\bibliography{reference}
\end{small}
\onecolumn
\appendix
\setcounter{secnumdepth}{1}
\input{sec/appendix}
\end{document}

%% file: sec/abstract.tex
Client heterogeneity creates both an opportunity and a risk in personalized federated graph learning. Knowledge held by other subgraphs may complement a receiver's Local model, but an incompatible transfer can override reliable predictions. One-shot communication sharpens this tension because an unsuitable server return cannot be corrected later. We introduce \method, which treats collaborative knowledge as a compact correction to a complete Local predictor rather than as its replacement. Each client uploads a rank-$r$ update carrier and a diagonal sketch of propagated message moments. The server uses them to construct a propagation-aware, receiver-anchored correction, while the receiver retains its full Local model. Convex negative-log-likelihood calibration (CNLL) then selects one coefficient between Local and External logits using validation nodes; model parameters remain fixed and no feedback is sent. At Rank-6, personalized returns occupy 9.6--17.6\% of dense tensor bytes across the six evaluated datasets. The correction receives nonzero weight and improves both Accuracy and weighted-F1 over Local on five datasets; on ogbn-arxiv, CNLL assigns zero predictive weight to the correction and preserves Local predictions exactly. Applying the same CNLL rule to matched baselines on three citation datasets does not account for these gains. The central result is therefore that a small transported correction can augment a complete Local model when receiver evidence supports it while leaving the Local prediction unchanged otherwise.

%% file: sec/1_introduction.tex
\section{Introduction}

\begin{figure}[!t]
\centering
\includegraphics[width=0.87\columnwidth]{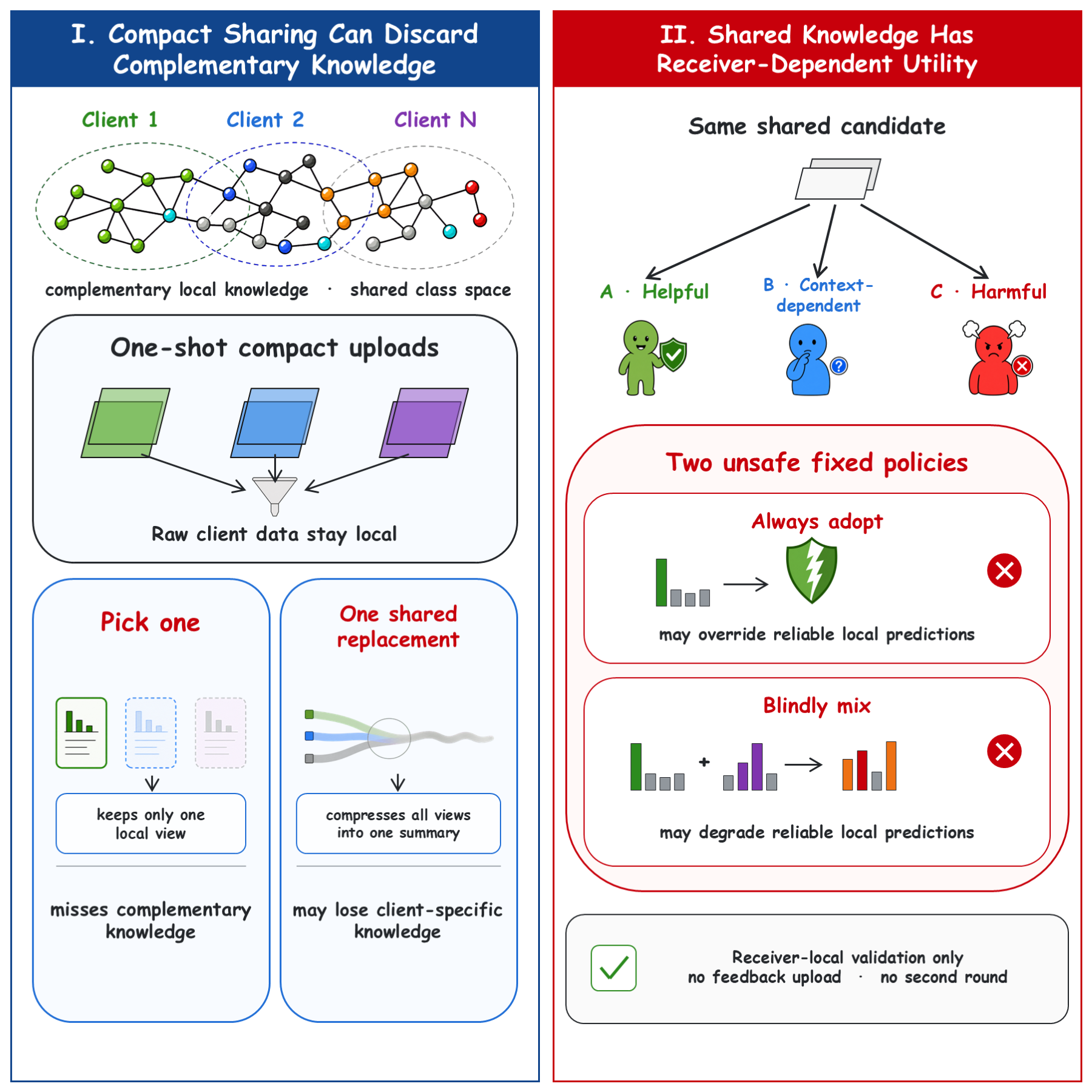}
\caption{Two coupled challenges in one-shot personalized FGL. Compact communication must preserve complementary evidence across clients, while the utility of shared knowledge depends on the receiver and can become negative under mismatch.}
\label{fig:motivation_challenges}
\end{figure}

Graph data are often distributed as interrelated local subgraphs: organizations or devices observe distinct communities, while raw nodes, edges, and labels cannot be pooled. Federated learning enables these clients to collaborate without collecting their records in one place \cite{mcmahan2017communication}. Structural diversity across subgraphs creates an opportunity because one community may contain evidence missing from another. It also creates a risk: a single global model can collapse incompatible knowledge and erase information that a local GNN already represents well \cite{zhang2021subgraph,baek2023personalized}. Personalized subgraph FL must obtain the former benefit without incurring the latter cost.

One-shot communication leaves no later round in which to repair a poor collaboration. The server must therefore return enough information to be useful, yet each receiver must remain able to limit its influence. FAFI traces a central failure mode of one-shot FL to inconsistent local representations and predictions \cite{zeng2025fafi}. Graph-specific one-shot methods instead construct a server surrogate graph, consolidate proxy models, or synthesize structural support \cite{yan2024opfgl,qian2025ghost,wan2025oasis}. Personalized FL can also raise average performance while harming individual clients \cite{wu2023fedora}. Figure~\ref{fig:motivation_challenges} summarizes the resulting tension between preserving complementary evidence during compact communication and preventing negative transfer at the receiver.

Existing approaches do not yet produce the deployment studied here. Multi-round personalization relies on later interaction, whereas many one-shot methods synthesize server data, optimize after aggregation, or return a consolidated replacement model. A compressed update also ignores how GCN propagation changes its parameter relevance across subgraphs. Even a well-constructed collaborative candidate may be unsuitable for a receiver with a strong Local predictor. The unresolved design problem is to transport graph-aware information compactly, anchor it to the receiver's complete Local state, and let that receiver control its contribution with no further model update or message.

\method separates compact transport from receiver adoption. Its primary contribution is a propagation-aware low-rank correction that adds external knowledge to the complete Local model instead of replacing that model. PACE synthesizes no graph at the server and performs no model optimization after download. A uniform, self-inclusive RegMean consensus supplies shared information, while subtraction of the receiver's own carrier produces its correction. The receiver then minimizes validation NLL along the segment between Local and External logits. CNLL may admit the external contribution, attenuate it, or recover Local exactly through $\alpha=0$.

Our contributions are:
\begin{itemize}
\item We formulate external collaboration as a compact correction to every receiver's complete Local model. One low-rank upload and one personalized low-rank return carry the shared displacement; no dense replacement model, graph synthesis, subsequent optimization, or feedback is required.
\item We apply propagation-aware RegMean to Rank-6 update carriers and compact second moments of propagated messages. Each return contains only the receiver's displacement from the shared consensus.
\item We make the transported predictor optional through a receiver-local information-geometric gate. CNLL calibrates one convex logit coefficient on the validation mask, and $\alpha=0$ restores the exact Local prediction with no additional update or message.
\item We evaluate PACE across six graph benchmarks under a matched one-shot protocol. PACE achieves higher mean Accuracy and weighted-F1 than Local on five benchmarks; on ogbn-arxiv, CNLL assigns zero predictive weight to the correction and preserves Local predictions exactly. Rank, communication, and matched-calibration analyses characterize when the external correction is retained.
\end{itemize}

%% file: sec/2_related_work.tex
\section{Related Work}

\paragraph{Personalization under heterogeneous clients.}
FedProx and SCAFFOLD reduce client drift through regularization or control variates \cite{li2020fedprox,karimireddy2020scaffold}. Personalized FL goes further by modifying the local objective, dividing the model into shared and private parts, or changing the collaboration relation itself. pFedMe and Ditto use regularized personalized objectives \cite{dinh2020pfedme,li2021ditto}. Per-FedAvg learns an initialization for later adaptation, whereas FedRep shares a representation and retains client-specific heads \cite{fallah2020perfedavg,collins2021fedrep}. FedFomo, FedAMP, and pFedGraph estimate which clients should influence one another from model information \cite{zhang2021fedfomo,huang2021fedamp,ye2023pfedgraph}; FedBN keeps normalization local under feature shift \cite{li2021fedbn}. Graph-personalized methods further use structural information. FedEgo trains with ego-graph representations and adaptively mixes local and global personalization-layer weights for each client \cite{zhang2023fedego}. FED-PUB derives personalized aggregation weights from functional embeddings of local GNNs, while FedAux represents clients through learned auxiliary projections \cite{baek2023personalized,zhuo2025personalized}. SubPFed combines functional embeddings with structural similarity derived from overlapping-node degrees to weight client-specific aggregation \cite{li2026subpfed}. These approaches personalize model components, local--global mixtures, or donor relations during federated optimization. A separate adoption question remains because better average performance may still conceal harm to individual clients. FEDORA exposes this choice through client-specific selective regularization that can suppress harmful parameter propagation \cite{wu2023fedora}. PACE instead makes a post-training decision between one retained Local predictor and one returned External candidate: receiver-local calibration controls this fixed candidate after the one-shot return rather than learning an aggregation relation across rounds.

\paragraph{Federated graph learning.}
Graphs add structural heterogeneity to the statistical shifts already present in FL. GCN and GraphSAGE derive node representations through neighborhood propagation \cite{kipf2017gcn,hamilton2017graphsage}, while subgraph FL loses neighbors that cross client boundaries \cite{zhang2021subgraph}. FGL methods respond at several levels. GCFL clusters clients using GNN gradients and gradient sequences; FedStar separates shareable structural knowledge from private feature knowledge \cite{xie2021gcfl,tan2023fedstar}. FED-PUB learns personalized aggregation and local masks for distributed node classification \cite{baek2023personalized}. FedGTA uses topology-aware smoothing statistics, FGGP exchanges prototypes across graph domains, and FedTAD distills according to class reliability under node and topology variation \cite{li2024fedgta,wan2024federated,zhu2024fedtad}. FedPPD follows a server-synthesis route: clients provide local prototypes along with model parameters and label distributions; a prototype-guided generator constructs a pseudo graph, which supports data-free distillation into the aggregated global GNN \cite{lin2025fedppd}. These approaches make topology affect collaboration through aggregation statistics, prototypes, synthesized structural support, or server optimization. PACE instead operates in model space after local training, retains each completed Local model, and returns a compact correction without constructing a server graph.

\paragraph{One-shot federated consolidation.}
Removing repeated communication turns personalization into a consolidation problem: independently trained client knowledge must be combined in one server stage. One family distills predictions through a surrogate input space. FedDF uses public unlabeled data, DENSE generates inputs, and FedSD2C communicates synthetic distillates \cite{lin2020feddf,zhang2022dense,zhang2024fedsd2c}. A second family combines parameters using Fisher information or layerwise posterior approximations \cite{jhunjhunwala2024fedfisher,liu2024fedlpa}; FAFI first aligns inconsistent representations and prototypes \cite{zeng2025fafi}. Graph-specific methods additionally introduce structural surrogates or client collaboration relations. O-pFGL constructs a global surrogate graph from class-wise feature statistics and then performs two-stage personalized training \cite{yan2024opfgl}. GHOST integrates client proxy models while consolidating parameters identified as important to topology, whereas OASIS combines a synergy-graph synthesizer, a topological codebook, and server-side distillation to produce a generalizable global model \cite{qian2025ghost,wan2025oasis}. pFedGNN privately estimates a global Laplacian and derives a client-level collaboration graph through one-shot graph inference; the inferred edges then guide graph-aware personalized parameter aggregation \cite{kataria2025learning}. Its one-shot claim concerns collaboration-graph construction rather than a complete one-upload/one-return training protocol. PACE instead communicates a small receiver-anchored correction to the complete Local model: it constructs no server graph, does not replace the trained Local state, and lets each receiver calibrate the correction after the return.

\paragraph{Compact communication and model merging.}
Low-rank structure offers one way to reduce the state exchanged during learning. PowerSGD compresses distributed gradients, FedPara parameterizes compact federated models, and LoRA learns low-rank updates to a frozen model \cite{vogels2019powersgd,nam2022fedpara,hu2022lora}. PACE instead factorizes an update after local training and uses the selected rank as an empirical performance--communication operating point. Its server step also draws on model merging. Weight averaging can succeed within a compatible basin \cite{wortsman2022soups}; task arithmetic represents specialization as a displacement from a shared initialization \cite{ilharco2023taskarithmetic}. Fisher merging and RegMean weight parameters through importance or activation geometry \cite{matena2022fisher,jin2023regmean}, while TIES-Merging and DARE handle sign conflict or sparsify displacements \cite{yadav2023ties,yu2024dare}. PACE adapts these ideas to graph propagation by transporting compact displacements together with propagated activation moments. The resulting candidate is anchored to the receiver before it is evaluated.

\paragraph{Selective adoption and confidence.}
A collaborative candidate still requires an adoption rule. Large softmax values need not represent calibrated correctness \cite{guo2017calibration}; deep ensembles provide an uncertainty baseline \cite{lakshminarayanan2017ensembles}, and selective prediction or energy scores can support rejection \cite{geifman2019selectivenet,hendrycks2017baseline,liu2020energy}. PACE considers a more specific choice between one fixed Local model and one returned External model. Validation NLL selects a single coefficient along their logit segment. The coefficient controls the External contribution for the receiver as a whole; it is neither a nodewise probability of transfer correctness nor a guarantee of improvement on every test node.

PACE connects these strands: propagation shapes the compact correction, the complete Local model remains available, and validation decides whether the external contribution is used.

%% file: sec/3_problem.tex
\section{Problem Formulation}

There are $K$ clients. Client $i$ owns a community-structured subgraph $G_i=(V_i,E_i,X_i)$ and labels on disjoint training and validation masks, $V_i^{\mathrm{tr}}$ and $V_i^{\mathrm{val}}$. All clients share a GCN architecture and a public initialization $W_0$. Client $i$ trains locally for a fixed budget and retains the last checkpoint as its complete Local model $W_i^L$. Raw nodes, edges, features, and labels remain local.

The protocol permits one upload $U_i$ from every client and one personalized return $D_i$ from the server. The upload contains a Rank-$r$ carrier for the local model update and a compact propagation-moment sketch. The return contains a Rank-$r$ correction constructed for the receiver. Complete serialized upload and download sizes, including sketches and framing overhead, are measured against the dense model size. The server receives no raw graph record, and no client performs another model update after the return.

\paragraph{Transport and calibration.}
The client models contain complementary information, but their usefulness varies across receivers. A single consolidated model can discard client-specific information, whereas applying a collaborative correction at full strength can introduce negative transfer. PACE therefore separates two decisions. Transport constructs an External candidate $W_i^E$ around the complete Local model. Calibration then chooses how much the candidate should affect the receiver's logits:
\begin{equation}
z_{i,v}(\alpha_i)=(1-\alpha_i)z_{i,v}^{L}+\alpha_i z_{i,v}^{E},
\qquad \alpha_i\in[0,1].
\label{eq:logit_blend}
\end{equation}
The receiver may use its validation labels only to solve this one-dimensional calibration problem. Validation does not select a checkpoint, trigger early stopping, update model parameters, or create another message. Test labels are reserved for final evaluation.

Message passing may use features and edges from the transductive local graph, but labels are accessed only through their declared masks. The protocol reduces exchanged state but does not provide a formal privacy guarantee: model carriers and moment sketches can disclose information and require a separate privacy analysis.

%% file: sec/4_method.tex
\section{PACE: Transport and Calibrate}

\begin{figure*}[t]
\centering
\begin{overpic}[width=\textwidth]{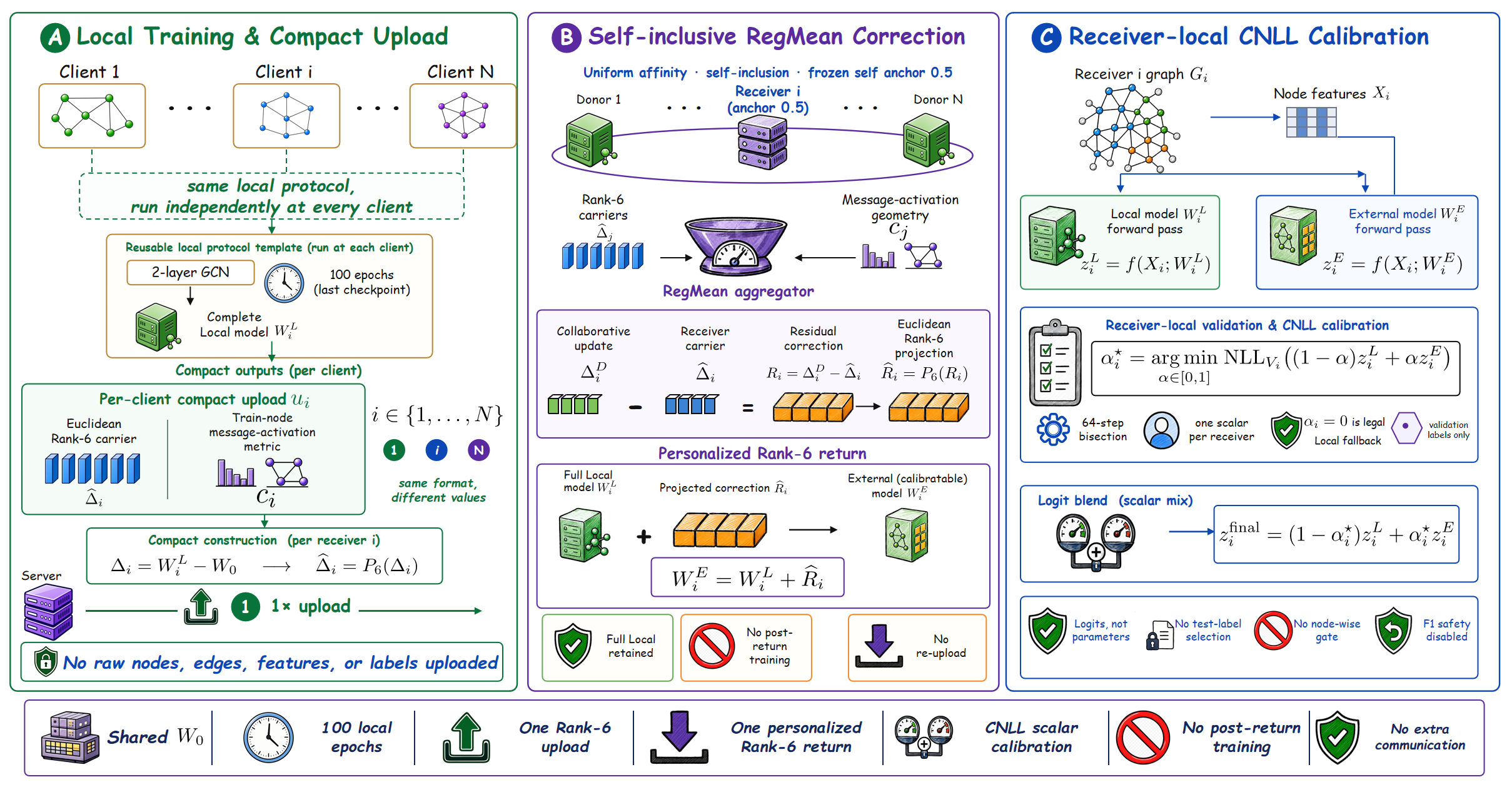}
\put(38.3,47.0){%
  \setlength{\fboxsep}{0pt}%
  \colorbox{white}{%
    \parbox[c][0.016\textwidth][c]{0.256\textwidth}{%
      \centering\color{FigureBlue}\tiny\bfseries
      Self-inclusive uniform weights 1/K}}}
\put(47.4,45.5){%
  \setlength{\fboxsep}{0pt}%
  \colorbox{white}{%
    \parbox[c][0.024\textwidth][c]{0.072\textwidth}{%
      \centering\color{FigureBlue}\tiny\bfseries
      Receiver i}}}
\end{overpic}
\caption{PACE pipeline. Each client trains locally and uploads a propagation-aware Rank-6 carrier once. The server forms a self-inclusive RegMean consensus and returns a receiver-anchored Rank-6 correction. Each receiver then calibrates a single convex logit coefficient with validation NLL, without parameter updates, test-label selection, feedback, or additional communication.}
\label{fig:pace_pipeline}
\end{figure*}

\method treats cross-client knowledge as a low-rank correction to a complete Local model, not as a replacement model. A frozen transport produces the correction through one compact upload and one personalized return; receiver-local CNLL then changes only its logit contribution and trains neither model. Figure~\ref{fig:pace_pipeline} gives the complete protocol.

\input{tab/main_results}

\input{tab/core_ablation}

\subsection{Local Training and Compact Upload}
After local training, client $i$ forms
\begin{equation}
\Delta_i=W_i^L-W_0,\qquad \bar\Delta_i=P_6(\Delta_i),
\label{eq:upload_carrier}
\end{equation}
where $P_6$ applies truncated SVD to matrix parameters. Vectors and other non-matrix parameters are transmitted without factorization. The wire representation stores low-rank factors rather than a reconstructed dense tensor.

For GCN layer $\ell$, let $H_i^{(\ell)}\in\mathbb{R}^{n_i\times d_\ell}$ be its input activation and $P_i$ the normalized propagation operator. We follow the PyG storage convention $W_i^{(\ell)}\in\mathbb{R}^{d_{\ell+1}\times d_\ell}$, under which the layer computes
\begin{equation}
H_i^{(\ell+1)}
=\sigma\!\left(P_iH_i^{(\ell)}
\left(W_i^{(\ell)}\right)^\top
+\mathbf{1}\left(b_i^{(\ell)}\right)^\top\right).
\end{equation}
The effective linear-layer input is therefore $S_i^{(\ell)}=P_iH_i^{(\ell)}\in\mathbb{R}^{n_i\times d_\ell}$. Client $i$ measures a coordinate-wise second moment on training center nodes,
\begin{equation}
m_i^{(\ell)}=\frac{1}{|V_i^{\mathrm{tr}}|}
\sum_{v\in V_i^{\mathrm{tr}}}
S_{i,v}^{(\ell)}\odot S_{i,v}^{(\ell)},
\end{equation}
and forms
\begin{equation}
\begin{aligned}
\widetilde m_i^{(\ell)}
&=\frac{m_i^{(\ell)}}{
\max(\operatorname{mean}(m_i^{(\ell)}),\epsilon)},\\
C_i^{(\ell)}
&=\operatorname{diag}(\widetilde m_i^{(\ell)}+\epsilon),
\qquad \epsilon=10^{-12}.
\end{aligned}
\end{equation}
Thus $m_i^{(\ell)}\in\mathbb{R}^{d_\ell}$ and $C_i^{(\ell)}\in\mathbb{R}^{d_\ell\times d_\ell}$ weight the input-coordinate columns of the stored parameter matrix. This convention makes the right multiplication in Eq.~\eqref{eq:regmean} dimensionally explicit. The layerwise normalization preserves relative propagation geometry without turning client-scale activation magnitude into an unintended donor weight. Sketch construction uses the training mask and does not read validation or test labels.

\subsection{Propagation-Aware Consensus}
The frozen configuration uses uniform, self-inclusive RegMean. Hence the collaborative consensus is shared across receivers. For a matrix parameter, the server solves
\begin{equation}
\Delta^C=
\arg\min_{\Delta}\sum_{j=1}^{K}
w_j\|(\Delta-\bar\Delta_j)C_j^{1/2}\|_F^2,
\qquad w_j=\frac{1}{K},
\label{eq:regmean}
\end{equation}
with closed form
\begin{equation}
\Delta^C=
\left(\sum_{j=1}^{K}w_j\bar\Delta_jC_j\right)
\left(\sum_{j=1}^{K}w_jC_j\right)^{-1}.
\end{equation}
Because $C_j$ is diagonal, the inverse is implemented by coordinate-wise division with every denominator clamped below by $\epsilon$. Biases, vectors, and other non-matrix parameters use the same uniform weighted mean. Rank-6 SVD is applied independently to each eligible two-dimensional matrix; matrices whose attainable rank is at most six and all non-matrix parameters remain dense. The consensus merges uploaded approximations, not unavailable dense updates.

\subsection{Receiver-Anchored Correction}
Returning a compressed consensus as a replacement would discard Local information outside the carrier subspace. PACE instead constructs
\begin{equation}
\Gamma_i=\Delta^C-\bar\Delta_i,\qquad
\widehat\Gamma_i=P_6(\Gamma_i),
\label{eq:correction}
\end{equation}
and the receiver obtains
\begin{equation}
W_i^E=W_i^L+\widehat\Gamma_i.
\end{equation}
The consensus itself is not personalized by receiver-dependent donor weights. Personalization arises from subtracting the receiver's carrier, applying the returned displacement to its complete Local state, and selecting its CNLL coefficient.

\subsection{Convex NLL Logit Calibration}
Let $z_{i,v}^{L}$ and $z_{i,v}^{E}$ denote the Local and External logits. The receiver selects one scalar
\begin{equation}
\alpha_i^\star=\arg\min_{\alpha\in[0,1]}
\operatorname{NLL}_{V_i^{\mathrm{val}}}\!\left(
(1-\alpha)z_i^{L}+\alpha z_i^{E}
\right).
\label{eq:cnll}
\end{equation}
CNLL is the computational realization of a receiver-local information-geometric gate: in probability space, the logit segment is a normalized geometric opinion pool and a weighted reverse-KL barycenter. Appendix~\ref{app:cnll_theory} proves this equivalence, objective convexity and moment matching, and a validation-NLL no-regret property. PACE checks the endpoint derivatives and otherwise uses 64 bisection iterations. If $V_i^{\mathrm{val}}$ is empty, the defined fallback is $\alpha_i^\star=0$. The deployed logits are $z_{i,v}(\alpha_i^\star)$ from Eq.~\eqref{eq:logit_blend}. The coefficient is not chosen by dataset name, client count, seed, or test performance.

The complete protocol therefore has one upload, one server return, zero post-return parameter updates, and zero feedback uploads.

%% file: tab/main_results.tex
\begin{table*}[t]
\centering
\small
\setlength{\tabcolsep}{4pt}
\renewcommand{\arraystretch}{0.92}
\begin{tabularx}{\textwidth}{l*{6}{Y}}
\toprule
\rowcolor{TableHeader}
Methods & \multicolumn{2}{c}{\textbf{Cora}} & \multicolumn{2}{c}{\textbf{CiteSeer}} & \multicolumn{2}{c}{\textbf{PubMed}} \\
\rowcolor{TableHeader}
 & Accuracy & W-F1 & Accuracy & W-F1 & Accuracy & W-F1 \\
\midrule
Local & \mbox{\underline{79.16$\pm$0.25}} & \mbox{\underline{79.08$\pm$0.27}} & \mbox{65.41$\pm$0.72} & \mbox{64.70$\pm$0.73} & \mbox{\underline{84.25$\pm$0.05}} & \mbox{\underline{84.19$\pm$0.05}} \\
\midrule
FedAvg & \mbox{33.18$\pm$0.94} & \mbox{20.28$\pm$1.74} & \mbox{72.79$\pm$0.32} & \mbox{69.81$\pm$0.46} & \mbox{78.21$\pm$1.07} & \mbox{76.49$\pm$1.49} \\
\rowcolor{TableStripe} FedProx & \mbox{39.61$\pm$2.45} & \mbox{30.88$\pm$3.61} & \mbox{\underline{72.82$\pm$0.32}} & \mbox{\underline{70.12$\pm$0.34}} & \mbox{72.32$\pm$2.21} & \mbox{68.28$\pm$2.37} \\
FedNova & \mbox{29.84$\pm$0.47} & \mbox{14.15$\pm$0.62} & \mbox{38.33$\pm$2.04} & \mbox{33.11$\pm$3.43} & \mbox{38.33$\pm$3.44} & \mbox{29.41$\pm$5.92} \\
\rowcolor{TableStripe} FedRCL & \mbox{23.57$\pm$8.89} & \mbox{15.44$\pm$6.53} & \mbox{26.31$\pm$3.14} & \mbox{18.02$\pm$3.37} & \mbox{35.70$\pm$8.32} & \mbox{19.27$\pm$6.76} \\
FedPub & \mbox{77.38$\pm$0.76} & \mbox{77.08$\pm$0.80} & \mbox{69.83$\pm$0.94} & \mbox{68.69$\pm$0.98} & \mbox{81.38$\pm$1.80} & \mbox{81.24$\pm$1.88} \\
\rowcolor{TableStripe} FedTAD & \mbox{33.73$\pm$0.79} & \mbox{21.25$\pm$1.35} & \mbox{72.58$\pm$0.25} & \mbox{69.56$\pm$0.34} & \mbox{78.96$\pm$1.10} & \mbox{77.22$\pm$1.60} \\
FedGTA & \mbox{44.31$\pm$1.27} & \mbox{36.24$\pm$1.99} & \mbox{71.22$\pm$0.30} & \mbox{68.16$\pm$0.25} & \mbox{62.10$\pm$1.89} & \mbox{59.19$\pm$2.51} \\
\rowcolor{TableStripe} FedAux-1R & \mbox{65.49$\pm$7.22} & \mbox{62.18$\pm$9.27} & \mbox{68.93$\pm$0.96} & \mbox{67.01$\pm$1.40} & \mbox{61.67$\pm$9.58} & \mbox{57.94$\pm$12.38} \\
\midrule
\rowcolor{PaceHighlight} \textbf{PACE (Ours)} & \mbox{\textbf{80.34$\pm$0.61}} & \mbox{\textbf{80.18$\pm$0.60}} & \mbox{\textbf{73.82$\pm$0.15}} & \mbox{\textbf{72.15$\pm$0.19}} & \mbox{\textbf{84.79$\pm$0.09}} & \mbox{\textbf{84.73$\pm$0.09}} \\
\addlinespace[1.5pt]
\midrule
\rowcolor{TableHeader}
Methods & \multicolumn{2}{c}{\textbf{CS}} & \multicolumn{2}{c}{\textbf{Computers}} & \multicolumn{2}{c}{\textbf{ogbn-arxiv}} \\
\rowcolor{TableHeader}
 & Accuracy & W-F1 & Accuracy & W-F1 & Accuracy & W-F1 \\
\midrule
Local & \mbox{\underline{89.17$\pm$0.10}} & \mbox{\underline{89.14$\pm$0.09}} & \mbox{\underline{87.87$\pm$0.32}} & \mbox{\underline{87.73$\pm$0.37}} & \mbox{\textbf{66.86$\pm$0.18}} & \mbox{\textbf{64.66$\pm$0.22}} \\
\midrule
FedAvg & \mbox{75.35$\pm$1.71} & \mbox{71.19$\pm$1.78} & \mbox{36.86$\pm$6.23} & \mbox{25.46$\pm$4.25} & \mbox{32.32$\pm$4.06} & \mbox{26.24$\pm$2.97} \\
\rowcolor{TableStripe} FedProx & \mbox{77.08$\pm$0.58} & \mbox{73.24$\pm$0.58} & \mbox{44.95$\pm$11.34} & \mbox{36.12$\pm$9.54} & \mbox{37.07$\pm$1.32} & \mbox{29.38$\pm$1.23} \\
FedNova & \mbox{45.76$\pm$8.30} & \mbox{35.68$\pm$8.59} & \mbox{37.13$\pm$0.13} & \mbox{20.54$\pm$0.25} & \mbox{14.24$\pm$3.41} & \mbox{5.24$\pm$1.67} \\
\rowcolor{TableStripe} FedRCL & \mbox{11.97$\pm$10.70} & \mbox{5.59$\pm$7.00} & \mbox{39.23$\pm$2.97} & \mbox{23.75$\pm$4.72} & \mbox{7.47$\pm$5.24} & \mbox{1.48$\pm$1.37} \\
FedPub & \mbox{89.00$\pm$0.23} & \mbox{88.92$\pm$0.25} & \mbox{86.75$\pm$0.49} & \mbox{86.49$\pm$0.72} & \mbox{\underline{60.00$\pm$0.24}} & \mbox{\underline{55.90$\pm$0.26}} \\
\rowcolor{TableStripe} FedTAD & \mbox{74.99$\pm$1.83} & \mbox{70.77$\pm$1.90} & \mbox{42.05$\pm$3.73} & \mbox{29.17$\pm$5.34} & \mbox{31.63$\pm$6.02} & \mbox{24.31$\pm$4.51} \\
FedGTA & \mbox{83.94$\pm$0.62} & \mbox{82.52$\pm$0.63} & \mbox{57.06$\pm$8.78} & \mbox{51.01$\pm$8.33} & \mbox{44.34$\pm$0.30} & \mbox{35.16$\pm$0.36} \\
\rowcolor{TableStripe} FedAux-1R & \mbox{77.97$\pm$7.82} & \mbox{74.93$\pm$9.42} & \mbox{78.55$\pm$8.61} & \mbox{76.37$\pm$10.64} & \mbox{56.80$\pm$2.92} & \mbox{52.69$\pm$2.82} \\
\midrule
\rowcolor{PaceHighlight} \textbf{PACE (Ours)} & \mbox{\textbf{89.40$\pm$0.16}} & \mbox{\textbf{89.30$\pm$0.17}} & \mbox{\textbf{87.99$\pm$0.23}} & \mbox{\textbf{87.84$\pm$0.30}} & \mbox{\textbf{66.86$\pm$0.18}} & \mbox{\textbf{64.66$\pm$0.22}} \\
\bottomrule
\end{tabularx}
\caption{Results with 10 clients under Louvain partitioning. Values are five-seed mean$\pm$sample standard deviation in percent. Bold and underline denote the best and second-best distinct displayed means, respectively.}
\label{tab:main_results}
\end{table*}

%% file: tab/core_ablation.tex
\begin{table*}[t]
\centering
\small
\setlength{\tabcolsep}{8pt}
\renewcommand{\arraystretch}{1.1}
\begin{tabularx}{\textwidth}{l*{3}{Y}}
\toprule
\rowcolor{TableHeader}
Variant & Cora & CiteSeer & PubMed \\
\midrule
Local & 79.16 & 65.41 & 84.25 \\
Transport ($\alpha=1$) & 63.46 & \textbf{73.82} & 83.66 \\
Fixed $\alpha=0.5$ & 80.09 & 69.96 & \textbf{84.97} \\
Mean + CNLL & 79.19 & 71.77 & 84.37 \\
Raw-Activation RegMean + CNLL & \underline{80.32} & 73.38 & 84.79 \\
Direct Return + CNLL & 79.91 & \underline{73.58} & \underline{84.80} \\
\rowcolor{PaceHighlight} \textbf{Full PACE} & \textbf{80.34} & \textbf{73.82} & 84.79 \\
\bottomrule
\end{tabularx}
\caption{Component ablation at $C=10$ over five seeds on the three citation benchmarks. Values are mean Accuracy (\%). Bold and underline denote the best and second-best distinct displayed means, respectively.}
\label{tab:core_ablation}
\end{table*}

%% file: sec/5_experiments.tex
\section{Experiments}

\subsection{Experimental Setup}
We evaluate PACE under Louvain community partitioning on Cora, CiteSeer, PubMed, CS, Computers, and ogbn-arxiv with 10 clients \cite{blondel2008louvain}. These six established subgraph-FGL benchmarks span citation, coauthor, product co-purchase, and large-scale OGB graphs under reproducible partitions and matched one-round protocols. The local train/validation/test proportions are approximately 20\%/40\%/40\% on Cora, CiteSeer, PubMed, CS, and Computers, and 60\%/20\%/20\% on ogbn-arxiv. Each client trains a two-layer GCN \cite{kipf2017gcn} with hidden width 64, dropout 0.5, Adam learning rate 0.01, weight decay 0.0005, and 100 local epochs from the same initialization; the last checkpoint is used. We report pooled test-node Accuracy and weighted-F1 as mean $\pm$ sample standard deviation over seeds 104729, 130363, 155921, 181081, and 206639.

The comparison includes independent Local training and eight collaborative methods: FedAvg, FedProx, FedNova, FedRCL, FedPub, FedTAD, FedGTA, and FedAux-1R \cite{mcmahan2017communication,li2020fedprox,wang2020fednova,seo2024fedrcl,baek2023personalized,zhu2024fedtad,li2024fedgta,zhuo2025personalized}. FedRCL uses its released relaxed-contrastive loss through a GCN layer-feature adapter. FedAux-1R is a one-round schedule adaptation that exposes its first personalized aggregate without post-return training. Every collaborative control receives one communication round and uses matched client partitions, initialization, architecture, and local training budget; the reported values therefore characterize matched one-round adaptations, not the methods' native multi-round convergence. We discuss O-pFGL as the closest personalized graph-specific one-shot formulation but do not report a self-reimplementation: its arXiv record at the July 2026 artifact freeze did not link official code, and reproducing its surrogate-graph construction and personalized training would introduce implementation-dependent differences \cite{yan2024opfgl}. PACE uses the frozen Rank-6 self-inclusive RegMean correction and CNLL in Eq.~\eqref{eq:cnll}. Validation labels select only $\alpha_i^\star$; they do not select checkpoints or update model parameters. Test labels are used only for evaluation.

\FloatBarrier
\subsection{Main Results}

The primary question in Table~\ref{tab:main_results} is whether a compact returned correction can augment or preserve the complete Local predictor. PACE assigns nonzero External weight and improves both Accuracy and weighted-F1 over Local on Cora, CiteSeer, PubMed, CS, and Computers; on ogbn-arxiv, every receiver selects $\alpha=0$ and reproduces Local. Among the displayed matched controls, PACE also has the highest mean of both metrics on those five datasets and ties Local on ogbn-arxiv. This ranking is supporting evidence for the correction design within the evaluated protocol, not the paper's primary contribution or a claim of universal dominance.

Appendix~\ref{app:receiver_diagnostics} reports seed-paired PACE-minus-Local differences, 95\% confidence intervals, and Win/Tie/Loss counts for both primary metrics. The intervals exclude zero on Cora, CiteSeer, and PubMed; CS and Computers retain positive mean differences with intervals crossing zero, while ogbn-arxiv is an exact tie. A post-selection diagnostic on the citation subset further records 118 helped, 15 tied, and 17 harmed receiver--seed units. Twenty-seven of 30 fixed receivers have nonnegative five-seed means; the worst receiver averages $-1.43$ points (worst single run: $-3.64$), precluding a worst-client safety claim.

The amount of external knowledge used is deliberately secondary to Local reliability. Receiver validation determines whether the complete Local predictor remains unchanged or admits an External contribution. We call $\alpha=0$ Local preservation, $0<\alpha<1$ controlled adoption, and $\alpha=1$ full adoption. These terms describe resolver behavior on validation data, not receiver-level test gains. Five benchmark settings show nonzero aggregate adoption together with improvements over Local on both primary metrics. Every ogbn-arxiv receiver instead selects $\alpha=0$ and reproduces Local exactly.

\begin{figure*}[!t]
  \centering
  \includegraphics[width=\textwidth]{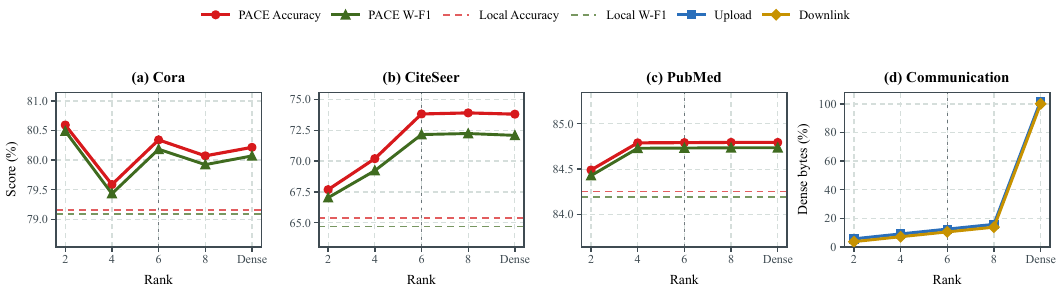}
  \caption{Rank sensitivity and communication at $C=10$. Dashed lines mark Local; the dotted line marks frozen Rank-6; Dense is the uncompressed control.}
  \label{fig:rank_tradeoff}
\end{figure*}

\subsection{Efficiency and Additional Analyses}
\label{sec:rank_tradeoff}
Figure~\ref{fig:rank_tradeoff} traces predictive performance and serialized communication as Rank changes. Rank-6 was fixed in the frozen protocol before the five-seed matrix continuation and before this diagnostic sweep; the sweep does not select a rank per dataset. Across the three rank-study benchmarks, Rank-6 averages 79.65\% Accuracy with 12.38\% upload and 10.47\% personalized-return bytes relative to dense tensors. Across all six datasets, its personalized returns occupy 9.6--17.6\% of dense tensor bytes (Table~\ref{tab:communication}), supporting the central use of a small correction while the complete Local model remains resident.

Under this fixed one-shot compact communication, CNLL does not force a receiver to use an unsupported correction: on ogbn-arxiv, it assigns zero predictive weight to the return and preserves the exact Local predictor without feedback communication or post-return model updates. Table~\ref{tab:communication} transparently reports the complete serialized traffic together with CNLL utilization on all six datasets.

\input{tab/appendix_tables}

\subsection{Transport and Resolver Analysis}

On the three citation datasets, Table~\ref{tab:core_ablation} separates the resolver, merging geometry, moment construction, and returned form. Relative to full adoption ($\alpha=1$), CNLL recovers 16.88 Accuracy points on Cora and 1.13 points on PubMed while retaining the already suitable CiteSeer candidate. A fixed $\alpha=0.5$ is competitive on Cora and slightly better on PubMed, but falls 3.86 points behind Full PACE on CiteSeer. The result supports receiver-adaptive adoption across datasets rather than a universally optimal fixed coefficient. Appendix~C extends this resolver diagnostic to all six datasets.

Holding the CNLL resolver and transported carriers fixed, replacing ordinary Mean with raw-activation RegMean improves Accuracy by 1.13, 1.61, and 0.42 points on Cora, CiteSeer, and PubMed, respectively, and wins all 15 paired dataset--seed comparisons. Replacing raw-activation moments with propagated-message moments adds a further 0.44 points on CiteSeer while matching the raw-moment variant on Cora and PubMed. Thus, RegMean geometry provides the most consistent component gain, whereas the additional benefit of propagation-aware moments is concentrated on CiteSeer.

Finally, returning a correction anchored to the complete Local model improves over direct consensus return by 0.43 and 0.24 points on Cora and CiteSeer and is effectively tied on PubMed. Across the 15 paired dataset--seed comparisons, Full PACE records 11 wins, one tie, and three losses. This pattern supports the receiver-anchored return as a modest but consistent refinement rather than the sole source of the overall gain.

\subsection{Matched Calibration Fairness}

We next apply the same CNLL calibration used by PACE to every baseline and report the highest-scoring matched control for each dataset and metric. As summarized in Table~\ref{tab:fairness_summary}, PACE remains ahead on both metrics for all three controlled citation datasets. Within this controlled subset, scalar calibration alone therefore does not reproduce PACE's result. Appendix~D contains the full native and Native+CNLL matrix.

\input{tab/fairness_summary}

Across the six $C=10$ settings, nonzero aggregate external influence coincides with gains over Local on both metrics, while ogbn-arxiv assigns zero predictive weight to the correction and preserves Local predictions exactly. PACE also exceeds the matched calibrated controls on all three citation datasets. CNLL uses neither test labels, model updates, nor feedback.

At $C=20/30$, PACE leads Local on both metrics for four of five reported datasets; Local remains stronger on Computers. Complete matrices and supporting ablations appear in Appendix~E.

\subsection{Limitations}
Louvain communities are a controlled proxy and may not represent other deployment partitions. PACE communicates trained-model statistics without a formal privacy guarantee and evaluates Local and External predictions before interpolation. Receiver-level harm diagnostics cover only the three citation datasets and do not establish a worst-client guarantee. Shared initialization and architecture encourage coordinate compatibility, but PACE does not resolve permutation symmetries. Rank-6 remains specific to this backbone and serialization scheme.

%% file: tab/appendix_tables.tex
\begin{table}[h]
\centering
\small
\setlength{\tabcolsep}{2pt}
\begin{tabularx}{\columnwidth}{l*{4}{Y}}
\toprule
\rowcolor{TableHeader}
Dataset & Mean $\alpha$ & $\alpha=0$ & Upload & Downlink \\
\midrule
Cora & 0.655 & 0.0\% & 12.1\% & 10.3\% \\
\rowcolor{TableStripe} CiteSeer & 0.979 & 0.0\% & 11.4\% & 9.7\% \\
PubMed & 0.278 & 0.0\% & 13.6\% & 11.4\% \\
\rowcolor{TableStripe} CS & 0.512 & 0.0\% & 11.2\% & 9.6\% \\
Computers & 0.134 & 22.0\% & 13.1\% & 11.1\% \\
\rowcolor{TableStripe} ogbn-arxiv & 0.000 & 100.0\% & 21.6\% & 17.6\% \\
\bottomrule
\end{tabularx}
\caption{PACE communication and CNLL utilization at $C=10$. Ratios use complete serialized Rank-6 PACE upload/download bytes relative to dense tensor bytes.}
\label{tab:communication}
\end{table}

%% file: tab/fairness_summary.tex
\begin{table}[ht]
\centering
\small
\setlength{\tabcolsep}{2.2pt}
\begin{tabularx}{\columnwidth}{ll*{3}{Y}}
\toprule
\rowcolor{TableHeader}
\multicolumn{5}{c}{\textbf{Accuracy}} \\
\rowcolor{TableHeader}
Dataset & Matched method & Score & PACE & Gap \\
\midrule
Cora & FedAux-1R+CNLL & 79.69 & \cellcolor{PaceHighlight}\textbf{80.34} & +0.65 \\
\rowcolor{TableStripe} CiteSeer & FedAvg+CNLL & 73.11 & \cellcolor{PaceHighlight}\textbf{73.82} & +0.71 \\
PubMed & FedTAD+CNLL & 84.44 & \cellcolor{PaceHighlight}\textbf{84.79} & +0.35 \\
\addlinespace[2pt]
\rowcolor{TableHeader}
\multicolumn{5}{c}{\textbf{Weighted-F1}} \\
\rowcolor{TableHeader}
Dataset & Matched method & Score & PACE & Gap \\
\midrule
Cora & FedAux-1R+CNLL & 79.64 & \cellcolor{PaceHighlight}\textbf{80.18} & +0.54 \\
\rowcolor{TableStripe} CiteSeer & FedProx+CNLL & 71.65 & \cellcolor{PaceHighlight}\textbf{72.15} & +0.50 \\
PubMed & FedTAD+CNLL & 84.38 & \cellcolor{PaceHighlight}\textbf{84.73} & +0.36 \\
\bottomrule
\end{tabularx}
\caption{Matched-CNLL fairness summary at $C=10$ over five seeds. For each metric and dataset, the matched method is the highest-scoring baseline after receiving the same receiver-local CNLL calibration; Gap is PACE-CNLL minus its score in percentage points.}
\label{tab:fairness_summary}
\end{table}

%% file: sec/6_conclusion.tex
\section{Conclusion}
PACE shows that cross-client knowledge can augment, rather than replace, a complete Local predictor through a compact propagation-aware correction. Rank-6 returns occupy 9.6--17.6\% of dense tensor bytes. CNLL assigns nonzero weight on five datasets and zero on ogbn-arxiv, exactly preserving Local predictions. Matched controls indicate that CNLL alone does not explain the citation gains. PACE's main contribution is this receiver-dependent compact-correction interface, not universal performance dominance.

%% file: sec/appendix.tex
\begin{multicols}{2}
\section{Seed-Paired and Receiver-Level Diagnostics}
\label{app:receiver_diagnostics}

Figure~\ref{fig:supplement_transfer_diagnostics}(a--b) pairs PACE and Local within the same dataset, partition seed, initialization, and client count. The intervals use the five seed-level differences rather than independent method summaries or node-level observations. They exclude zero on Cora, CiteSeer, and PubMed; CS and Computers have positive means with intervals crossing zero, and ogbn-arxiv is an exact displayed tie.

Panel (c) unpools the citation results into receiver--seed units. Each $\alpha_i^\star$ is fixed using receiver-local validation labels before Local and PACE are compared on that receiver's test nodes. Across 150 units, 118 improve, 15 tie, and 17 decline. Averaging each fixed receiver over five training seeds leaves 27 of 30 receiver means nonnegative. The worst receiver mean is $-1.43$ Accuracy points and the worst single unit is $-3.64$ points. Test labels enter only this post-selection report, which measures observed heterogeneity rather than establishing a per-client safety guarantee.

\end{multicols}

\begin{center}
  \centering
  \includegraphics[width=\textwidth]{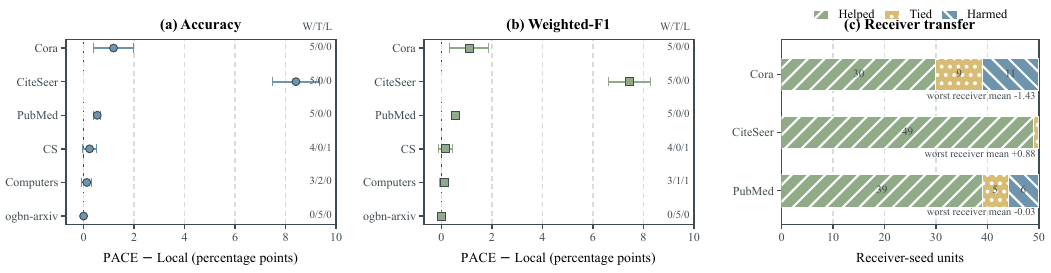}
  \captionof{figure}{Transfer diagnostics at $C=10$. Panels (a--b) show mean PACE-minus-Local differences with 95\% paired $t$ intervals over five matched seeds; right-side labels are Win/Tie/Loss counts at a 0.01-point tolerance. Panel (c) shows helped, tied, and harmed receiver--seed units for the citation datasets; parenthetical labels give the worst fixed-receiver mean over five seeds.}
  \label{fig:supplement_transfer_diagnostics}
\end{center}

\begin{minipage}[t]{0.31\textwidth}
\vspace{0pt}
\section{Exact Rank Study}

Table~\ref{tab:rank_tradeoff} reports the exact values underlying the main
paper's Rank-sensitivity figure. Rank 6 is fixed globally rather than selected
per dataset. The same choice is used for the low-rank upload and personalized
return throughout the reported experiments.
\end{minipage}\hfill
\begin{minipage}[t]{0.66\textwidth}
\vspace{0pt}
\input{tab/rank_tradeoff}
\end{minipage}

\clearpage
\section{Complete Resolver Ablation}
\label{app:resolver_ablation}

Table~\ref{tab:extended_ablation} extends the main-paper resolver ablation to
every evaluated dataset--client setting using the final five-seed summaries.
It holds the transported correction fixed and changes only receiver-local CNLL
calibration.

\input{tab/extended_ablation}

\section{Matched CNLL Fairness Controls}
\label{app:fairness_controls}

Table~\ref{tab:fairness_cnll} applies the same receiver-local scalar CNLL calibration to every baseline using that method's own External predictor. These controls do not replace native baseline identities in the main table; they test whether PACE's result can be explained by calibration alone.

\input{tab/fairness_cnll}

\clearpage
\input{tab/full_louvain_results}

\begin{minipage}[t]{0.48\textwidth}
\paragraph{Partition and evaluation protocol.}
All methods use the same audited OpenFGL partition and mask cache within each dataset--client setting. We run \texttt{subgraph\_fl\_louvain} with data seed 2024, Louvain resolution 1.0, and size-balancing tolerance $\delta=20$. Communities larger than the target client capacity are split before groups are assigned to clients; each client graph is then the induced subgraph on its assigned nodes, so cross-client edges are not retained. The train/validation/test proportions, model seeds, architecture, optimizer, and local budget match the main-paper setup. These choices remain fixed across Local, all collaborative baselines, and PACE.
\end{minipage}\hfill
\begin{minipage}[t]{0.48\textwidth}

\begin{center}
\large\bfseries Robustness to Graph Partitioning
\end{center}

Louvain is the canonical partition used in the main study. We additionally
evaluate METIS and Dirichlet label-skew partitions without changing the model,
optimization, client count, seeds, metrics, or one-round protocol. Both tables
cover Cora, CiteSeer, PubMed, CS, and Computers at $C=10$ and report five-seed
mean$\pm$sample standard deviation for the complete ten-method matrix.
No unavailable dataset--partition result is reconstructed.
\end{minipage}

\clearpage
\input{tab/partition_robustness}

\clearpage
\begin{multicols}{2}
\input{sec/cnll_theory}
\end{multicols}

%% file: tab/rank_tradeoff.tex
\centering
\small
\setlength{\tabcolsep}{2.6pt}
\begin{tabularx}{\linewidth}{l*{5}{Y}}
\toprule
\rowcolor{TableHeader}
Rank & Cora & CiteSeer & PubMed & Up. & Down. \\
\midrule
2 & 80.59 & 67.69 & 84.49 & 5.58\% & 3.66\% \\
\rowcolor{TableStripe} 4 & 79.59 & 70.20 & 84.79 & 9.01\% & 7.10\% \\
\rowcolor{PaceHighlight} \textbf{6} & 80.34 & 73.82 & 84.79 & 12.38\% & 10.47\% \\
\rowcolor{TableStripe} 8 & 80.07 & 73.91 & 84.79 & 15.69\% & 13.78\% \\
Dense & 80.22 & 73.81 & 84.79 & 101.95\% & 100.04\% \\
\bottomrule
\end{tabularx}
\captionof{table}{Five-seed Rank study at $C=10$ using the selected CNLL resolver. Accuracy is in percent; upload and personalized downlink ratios are relative to dense tensor bytes and averaged across the three datasets. Rank-6 is frozen globally.}
\label{tab:rank_tradeoff}
\par\medskip

%% file: tab/extended_ablation.tex
\centering
\small
\setlength{\tabcolsep}{4pt}
\begin{tabularx}{\textwidth}{lr*{5}{Y}}
\toprule
\rowcolor{TableHeader}
Dataset & $C$ & \multicolumn{2}{c}{Transport} & \multicolumn{2}{c}{Transport + CNLL} & $\Delta$ Acc. \\
\rowcolor{TableHeader}
& & Accuracy & W-F1 & Accuracy & W-F1 & (points) \\
\midrule
Cora & 10 & 63.46$\pm$1.40 & 61.31$\pm$1.65 & \cellcolor{PaceHighlight}80.34$\pm$0.61 & \cellcolor{PaceHighlight}80.18$\pm$0.60 & +16.88 \\
\rowcolor{TableStripe} Cora & 20 & 48.19$\pm$1.71 & 42.46$\pm$2.27 & \cellcolor{PaceHighlight}76.71$\pm$0.49 & \cellcolor{PaceHighlight}76.42$\pm$0.51 & +28.52 \\
Cora & 30 & 35.56$\pm$2.20 & 24.40$\pm$3.47 & \cellcolor{PaceHighlight}74.88$\pm$0.49 & \cellcolor{PaceHighlight}74.48$\pm$0.50 & +39.32 \\
\rowcolor{TableStripe} CiteSeer & 10 & 73.82$\pm$0.21 & 72.12$\pm$0.32 & \cellcolor{PaceHighlight}73.82$\pm$0.15 & \cellcolor{PaceHighlight}72.15$\pm$0.19 & +0.00 \\
CiteSeer & 20 & 69.02$\pm$1.27 & 67.63$\pm$1.16 & \cellcolor{PaceHighlight}69.36$\pm$0.87 & \cellcolor{PaceHighlight}68.11$\pm$0.89 & +0.33 \\
\rowcolor{TableStripe} CiteSeer & 30 & 68.29$\pm$0.57 & 65.65$\pm$0.54 & \cellcolor{PaceHighlight}68.69$\pm$0.59 & \cellcolor{PaceHighlight}66.31$\pm$0.63 & +0.40 \\
PubMed & 10 & 83.66$\pm$0.40 & 83.61$\pm$0.42 & \cellcolor{PaceHighlight}84.79$\pm$0.09 & \cellcolor{PaceHighlight}84.73$\pm$0.09 & +1.13 \\
\rowcolor{TableStripe} PubMed & 20 & 81.66$\pm$0.57 & 81.55$\pm$0.62 & \cellcolor{PaceHighlight}83.45$\pm$0.12 & \cellcolor{PaceHighlight}83.41$\pm$0.12 & +1.79 \\
PubMed & 30 & 81.21$\pm$0.56 & 81.17$\pm$0.62 & \cellcolor{PaceHighlight}83.75$\pm$0.08 & \cellcolor{PaceHighlight}83.72$\pm$0.09 & +2.55 \\
\rowcolor{TableStripe} CS & 10 & 82.68$\pm$0.99 & 81.17$\pm$1.22 & \cellcolor{PaceHighlight}89.40$\pm$0.16 & \cellcolor{PaceHighlight}89.30$\pm$0.17 & +6.72 \\
CS & 20 & 81.99$\pm$2.02 & 80.55$\pm$2.68 & \cellcolor{PaceHighlight}87.96$\pm$0.13 & \cellcolor{PaceHighlight}87.82$\pm$0.15 & +5.97 \\
\rowcolor{TableStripe} CS & 30 & 78.66$\pm$1.78 & 75.99$\pm$2.18 & \cellcolor{PaceHighlight}86.53$\pm$0.16 & \cellcolor{PaceHighlight}86.23$\pm$0.15 & +7.87 \\
Computers & 10 & 32.21$\pm$8.41 & 22.51$\pm$4.04 & \cellcolor{PaceHighlight}87.99$\pm$0.23 & \cellcolor{PaceHighlight}87.84$\pm$0.30 & +55.78 \\
\rowcolor{TableStripe} Computers & 20 & 35.01$\pm$3.58 & 25.45$\pm$1.72 & \cellcolor{PaceHighlight}86.34$\pm$0.57 & \cellcolor{PaceHighlight}85.91$\pm$0.82 & +51.33 \\
Computers & 30 & 31.41$\pm$6.86 & 21.13$\pm$2.55 & \cellcolor{PaceHighlight}85.33$\pm$0.12 & \cellcolor{PaceHighlight}85.10$\pm$0.11 & +53.92 \\
\rowcolor{TableStripe} ogbn-arxiv & 10 & 40.13$\pm$2.00 & 32.85$\pm$2.19 & \cellcolor{PaceHighlight}66.86$\pm$0.18 & \cellcolor{PaceHighlight}64.66$\pm$0.22 & +26.73 \\
ogbn-arxiv & 20 & 31.23$\pm$4.19 & 25.82$\pm$3.15 & \cellcolor{PaceHighlight}66.39$\pm$0.17 & \cellcolor{PaceHighlight}64.33$\pm$0.19 & +35.16 \\
\rowcolor{TableStripe} ogbn-arxiv & 30 & 28.57$\pm$1.96 & 22.69$\pm$2.37 & \cellcolor{PaceHighlight}64.93$\pm$0.13 & \cellcolor{PaceHighlight}62.89$\pm$0.14 & +36.36 \\
\bottomrule
\end{tabularx}
\captionof{table}{Complete shared-transport ablation over all evaluated dataset--client settings and five seeds. $\Delta$ Accuracy is Transport+CNLL minus Transport in percentage points.}
\label{tab:extended_ablation}

%% file: tab/fairness_cnll.tex
\centering
\scriptsize
\setlength{\tabcolsep}{3.2pt}
\renewcommand{\arraystretch}{0.80}
\begin{tabularx}{\textwidth}{ll*{6}{Y}}
\toprule
\rowcolor{TableHeader}
\textbf{Method} & \textbf{Calibration} & \multicolumn{2}{c}{\textbf{Cora}} & \multicolumn{2}{c}{\textbf{CiteSeer}} & \multicolumn{2}{c}{\textbf{PubMed}} \\
\rowcolor{TableHeader}
& & Accuracy & W-F1 & Accuracy & W-F1 & Accuracy & W-F1 \\
\midrule
Local & None & 79.16$\pm$0.25 & 79.08$\pm$0.27 & 65.41$\pm$0.72 & 64.70$\pm$0.73 & 84.25$\pm$0.05 & 84.19$\pm$0.05 \\
\rowcolor{TableStripe} Local & CNLL & 79.16$\pm$0.25 & 79.08$\pm$0.27 & 65.41$\pm$0.72 & 64.70$\pm$0.73 & 84.25$\pm$0.05 & 84.19$\pm$0.05 \\
\addlinespace[1pt]
FedAvg & None & 33.18$\pm$0.94 & 20.28$\pm$1.74 & 72.79$\pm$0.32 & 69.81$\pm$0.46 & 78.21$\pm$1.07 & 76.49$\pm$1.49 \\
\rowcolor{TableStripe} FedAvg & CNLL & 79.37$\pm$0.54 & 79.21$\pm$0.56 & 73.11$\pm$0.49 & 71.48$\pm$0.53 & 84.43$\pm$0.08 & 84.36$\pm$0.08 \\
\addlinespace[1pt]
FedProx & None & 39.61$\pm$2.45 & 30.88$\pm$3.61 & 72.82$\pm$0.32 & 70.12$\pm$0.34 & 72.32$\pm$2.21 & 68.28$\pm$2.37 \\
\rowcolor{TableStripe} FedProx & CNLL & 79.46$\pm$0.24 & 79.36$\pm$0.24 & 72.99$\pm$0.34 & 71.65$\pm$0.37 & 83.80$\pm$0.09 & 83.74$\pm$0.09 \\
\addlinespace[1pt]
FedNova & None & 29.84$\pm$0.47 & 14.15$\pm$0.62 & 38.33$\pm$2.04 & 33.11$\pm$3.43 & 38.33$\pm$3.44 & 29.41$\pm$5.92 \\
\rowcolor{TableStripe} FedNova & CNLL & 59.55$\pm$1.01 & 57.43$\pm$1.18 & 50.52$\pm$0.72 & 48.84$\pm$0.99 & 66.74$\pm$0.63 & 68.04$\pm$0.54 \\
\addlinespace[1pt]
FedRCL & None & 23.57$\pm$8.89 & 15.44$\pm$6.53 & 26.31$\pm$3.14 & 18.02$\pm$3.37 & 35.70$\pm$8.32 & 19.27$\pm$6.76 \\
\rowcolor{TableStripe} FedRCL & CNLL & 54.42$\pm$3.63 & 53.46$\pm$3.63 & 38.94$\pm$2.99 & 34.31$\pm$5.70 & 61.97$\pm$4.51 & 58.18$\pm$7.59 \\
\addlinespace[1pt]
FedPub & None & 77.38$\pm$0.76 & 77.08$\pm$0.80 & 69.83$\pm$0.94 & 68.69$\pm$0.98 & 81.38$\pm$1.80 & 81.24$\pm$1.88 \\
\rowcolor{TableStripe} FedPub & CNLL & 78.65$\pm$0.43 & 78.45$\pm$0.45 & 68.91$\pm$0.64 & 67.89$\pm$0.62 & 83.85$\pm$0.27 & 83.79$\pm$0.26 \\
\addlinespace[1pt]
FedTAD & None & 33.73$\pm$0.79 & 21.25$\pm$1.35 & 72.58$\pm$0.25 & 69.56$\pm$0.34 & 78.96$\pm$1.10 & 77.22$\pm$1.60 \\
\rowcolor{TableStripe} FedTAD & CNLL & 79.32$\pm$0.49 & 79.16$\pm$0.50 & 73.05$\pm$0.32 & 71.41$\pm$0.36 & 84.44$\pm$0.08 & 84.38$\pm$0.07 \\
\addlinespace[1pt]
FedGTA & None & 44.31$\pm$1.27 & 36.24$\pm$1.99 & 71.22$\pm$0.30 & 68.16$\pm$0.25 & 62.10$\pm$1.89 & 59.19$\pm$2.51 \\
\rowcolor{TableStripe} FedGTA & CNLL & 79.59$\pm$0.61 & 79.44$\pm$0.63 & 72.67$\pm$0.25 & 71.04$\pm$0.31 & 84.32$\pm$0.05 & 84.26$\pm$0.05 \\
\addlinespace[1pt]
FedAux-1R & None & 65.49$\pm$7.22 & 62.18$\pm$9.27 & 68.93$\pm$0.96 & 67.01$\pm$1.40 & 61.67$\pm$9.58 & 57.94$\pm$12.38 \\
\rowcolor{TableStripe} FedAux-1R & CNLL & 79.69$\pm$0.23 & 79.64$\pm$0.24 & 68.12$\pm$0.82 & 66.97$\pm$0.88 & 83.58$\pm$0.28 & 83.50$\pm$0.29 \\
\addlinespace[1pt]
\midrule
\rowcolor{PaceHighlight} \textbf{PACE} & \textbf{CNLL} & \textbf{80.34$\pm$0.61} & \textbf{80.18$\pm$0.60} & \textbf{73.82$\pm$0.15} & \textbf{72.15$\pm$0.19} & \textbf{84.79$\pm$0.09} & \textbf{84.73$\pm$0.09} \\
\bottomrule
\end{tabularx}
\captionof{table}{Matched calibration fairness controls at $C=10$ over five seeds. Each baseline is shown in its native form and with the same receiver-local CNLL logit calibration; PACE uses its Rank-6 transported External predictor. Values are mean$\pm$sample standard deviation in percent, with column best in bold.}
\label{tab:fairness_cnll}
\par\medskip

%% file: tab/full_louvain_results.tex
\section{Complete Louvain Results}
\label{app:full_louvain}
The main paper reports the complete $C=10$ Accuracy and weighted-F1 matrix. Tables~\ref{tab:full_primary_c20} and~\ref{tab:full_primary_c30} use the same layout for the additional client counts. Table~\ref{tab:full_macro_f1} collects fixed-class Macro-F1 for all settings. Values are five-seed mean $\pm$ sample standard deviation in percent.
\centering
\scriptsize
\setlength{\tabcolsep}{3pt}
\renewcommand{\arraystretch}{0.75}
\begin{tabularx}{\textwidth}{l*{6}{Y}}
\toprule
\rowcolor{TableHeader}
Methods & \multicolumn{2}{c}{\textbf{Cora}} & \multicolumn{2}{c}{\textbf{CiteSeer}} & \multicolumn{2}{c}{\textbf{PubMed}} \\
\rowcolor{TableHeader}
 & Accuracy & W-F1 & Accuracy & W-F1 & Accuracy & W-F1 \\
\midrule
Local & \mbox{\underline{75.69$\pm$0.34}} & \mbox{\underline{75.57$\pm$0.35}} & \mbox{60.64$\pm$0.33} & \mbox{60.38$\pm$0.33} & \mbox{\underline{82.38$\pm$0.09}} & \mbox{\underline{82.35$\pm$0.08}} \\
\midrule
FedAvg & \mbox{29.85$\pm$0.08} & \mbox{13.79$\pm$0.17} & \mbox{62.85$\pm$6.29} & \mbox{59.86$\pm$6.80} & \mbox{74.02$\pm$1.53} & \mbox{71.15$\pm$2.10} \\
\rowcolor{TableStripe} FedProx & \mbox{30.22$\pm$0.34} & \mbox{14.56$\pm$0.70} & \mbox{\underline{65.63$\pm$1.79}} & \mbox{\underline{62.76$\pm$1.89}} & \mbox{68.32$\pm$2.80} & \mbox{63.76$\pm$3.13} \\
FedNova & \mbox{29.97$\pm$0.22} & \mbox{14.73$\pm$0.54} & \mbox{34.38$\pm$1.04} & \mbox{28.72$\pm$2.35} & \mbox{38.31$\pm$3.10} & \mbox{29.12$\pm$5.78} \\
\rowcolor{TableStripe} FedRCL & \mbox{28.81$\pm$1.69} & \mbox{20.91$\pm$2.94} & \mbox{25.23$\pm$3.93} & \mbox{19.50$\pm$3.91} & \mbox{32.02$\pm$10.19} & \mbox{16.28$\pm$8.29} \\
FedPub & \mbox{72.13$\pm$1.33} & \mbox{71.68$\pm$1.44} & \mbox{61.94$\pm$0.59} & \mbox{61.04$\pm$0.59} & \mbox{78.35$\pm$1.64} & \mbox{78.22$\pm$1.69} \\
\rowcolor{TableStripe} FedTAD & \mbox{29.96$\pm$0.18} & \mbox{14.01$\pm$0.38} & \mbox{62.81$\pm$6.41} & \mbox{59.61$\pm$7.04} & \mbox{75.73$\pm$1.08} & \mbox{73.14$\pm$1.56} \\
FedGTA & \mbox{52.47$\pm$0.28} & \mbox{47.52$\pm$0.50} & \mbox{56.38$\pm$6.01} & \mbox{52.70$\pm$7.19} & \mbox{74.82$\pm$2.04} & \mbox{73.14$\pm$2.49} \\
\rowcolor{TableStripe} FedAux-1R & \mbox{50.88$\pm$9.56} & \mbox{44.91$\pm$13.04} & \mbox{59.55$\pm$2.98} & \mbox{57.37$\pm$3.12} & \mbox{66.84$\pm$5.28} & \mbox{64.36$\pm$6.69} \\
\midrule
\rowcolor{PaceHighlight} \textbf{PACE (Ours)} & \mbox{\textbf{76.71$\pm$0.49}} & \mbox{\textbf{76.42$\pm$0.51}} & \mbox{\textbf{69.36$\pm$0.87}} & \mbox{\textbf{68.11$\pm$0.89}} & \mbox{\textbf{83.45$\pm$0.12}} & \mbox{\textbf{83.41$\pm$0.12}} \\
\bottomrule
\end{tabularx}
\vspace{2pt}
\begin{tabularx}{\textwidth}{l*{6}{Y}}
\toprule
\rowcolor{TableHeader}
Methods & \multicolumn{2}{c}{\textbf{CS}} & \multicolumn{2}{c}{\textbf{Computers}} & \multicolumn{2}{c}{\textbf{ogbn-arxiv}} \\
\rowcolor{TableHeader}
 & Accuracy & W-F1 & Accuracy & W-F1 & Accuracy & W-F1 \\
\midrule
Local & \mbox{\underline{86.95$\pm$0.11}} & \mbox{\underline{86.90$\pm$0.10}} & \mbox{\textbf{86.59$\pm$0.59}} & \mbox{\textbf{86.31$\pm$0.77}} & \mbox{\textbf{66.39$\pm$0.17}} & \mbox{\textbf{64.33$\pm$0.19}} \\
\midrule
FedAvg & \mbox{64.05$\pm$3.00} & \mbox{57.64$\pm$3.98} & \mbox{36.89$\pm$9.04} & \mbox{27.26$\pm$5.45} & \mbox{16.49$\pm$1.59} & \mbox{8.41$\pm$1.89} \\
\rowcolor{TableStripe} FedProx & \mbox{69.24$\pm$1.48} & \mbox{64.38$\pm$1.72} & \mbox{41.90$\pm$5.80} & \mbox{32.80$\pm$3.92} & \mbox{24.08$\pm$2.51} & \mbox{17.29$\pm$1.83} \\
FedNova & \mbox{46.77$\pm$7.63} & \mbox{37.71$\pm$7.74} & \mbox{36.88$\pm$0.17} & \mbox{20.43$\pm$0.11} & \mbox{13.35$\pm$3.56} & \mbox{4.26$\pm$1.54} \\
\rowcolor{TableStripe} FedRCL & \mbox{15.35$\pm$9.07} & \mbox{9.96$\pm$6.19} & \mbox{37.41$\pm$2.34} & \mbox{23.50$\pm$4.47} & \mbox{6.74$\pm$5.47} & \mbox{1.33$\pm$1.44} \\
FedPub & \mbox{85.69$\pm$0.42} & \mbox{85.48$\pm$0.48} & \mbox{84.18$\pm$0.48} & \mbox{83.25$\pm$0.69} & \mbox{58.69$\pm$0.88} & \mbox{54.66$\pm$1.30} \\
\rowcolor{TableStripe} FedTAD & \mbox{64.48$\pm$3.17} & \mbox{58.26$\pm$4.10} & \mbox{39.32$\pm$7.66} & \mbox{29.78$\pm$6.25} & \mbox{14.27$\pm$0.97} & \mbox{5.31$\pm$1.72} \\
FedGTA & \mbox{81.70$\pm$0.28} & \mbox{81.04$\pm$0.31} & \mbox{63.36$\pm$1.83} & \mbox{58.93$\pm$1.92} & \mbox{54.36$\pm$0.47} & \mbox{49.08$\pm$0.67} \\
\rowcolor{TableStripe} FedAux-1R & \mbox{64.29$\pm$7.25} & \mbox{59.11$\pm$9.44} & \mbox{66.24$\pm$12.22} & \mbox{61.69$\pm$15.55} & \mbox{53.45$\pm$6.17} & \mbox{50.61$\pm$5.68} \\
\midrule
\rowcolor{PaceHighlight} \textbf{PACE (Ours)} & \mbox{\textbf{87.96$\pm$0.13}} & \mbox{\textbf{87.82$\pm$0.15}} & \mbox{\underline{86.34$\pm$0.57}} & \mbox{\underline{85.91$\pm$0.82}} & \mbox{\textbf{66.39$\pm$0.17}} & \mbox{\textbf{64.33$\pm$0.19}} \\
\bottomrule
\end{tabularx}
\captionof{table}{Complete Louvain results for $C=20$. The layout and notation match the main-paper $C=10$ table.}
\label{tab:full_primary_c20}
\par\medskip

\centering
\scriptsize
\setlength{\tabcolsep}{3pt}
\renewcommand{\arraystretch}{0.75}
\begin{tabularx}{\textwidth}{l*{6}{Y}}
\toprule
\rowcolor{TableHeader}
Methods & \multicolumn{2}{c}{\textbf{Cora}} & \multicolumn{2}{c}{\textbf{CiteSeer}} & \multicolumn{2}{c}{\textbf{PubMed}} \\
\rowcolor{TableHeader}
 & Accuracy & W-F1 & Accuracy & W-F1 & Accuracy & W-F1 \\
\midrule
Local & \mbox{\underline{71.94$\pm$0.40}} & \mbox{\underline{71.81$\pm$0.40}} & \mbox{59.46$\pm$0.16} & \mbox{58.55$\pm$0.19} & \mbox{\underline{82.25$\pm$0.12}} & \mbox{\underline{82.21$\pm$0.12}} \\
\midrule
FedAvg & \mbox{29.50$\pm$0.00} & \mbox{13.44$\pm$0.00} & \mbox{62.71$\pm$4.61} & \mbox{59.96$\pm$4.14} & \mbox{70.59$\pm$0.95} & \mbox{66.97$\pm$1.23} \\
\rowcolor{TableStripe} FedProx & \mbox{29.50$\pm$0.00} & \mbox{13.44$\pm$0.00} & \mbox{\underline{64.84$\pm$1.78}} & \mbox{\underline{61.88$\pm$1.66}} & \mbox{65.41$\pm$2.56} & \mbox{59.99$\pm$2.44} \\
FedNova & \mbox{29.57$\pm$0.51} & \mbox{15.01$\pm$1.01} & \mbox{33.20$\pm$1.43} & \mbox{28.34$\pm$2.08} & \mbox{38.46$\pm$2.56} & \mbox{29.99$\pm$5.12} \\
\rowcolor{TableStripe} FedRCL & \mbox{27.31$\pm$3.03} & \mbox{20.10$\pm$2.03} & \mbox{27.19$\pm$2.86} & \mbox{23.10$\pm$2.93} & \mbox{35.68$\pm$8.29} & \mbox{19.25$\pm$6.74} \\
FedPub & \mbox{69.26$\pm$1.33} & \mbox{68.90$\pm$1.38} & \mbox{60.27$\pm$0.61} & \mbox{59.06$\pm$0.57} & \mbox{75.74$\pm$3.64} & \mbox{75.54$\pm$3.75} \\
\rowcolor{TableStripe} FedTAD & \mbox{29.50$\pm$0.00} & \mbox{13.44$\pm$0.00} & \mbox{61.26$\pm$6.29} & \mbox{58.24$\pm$6.21} & \mbox{73.00$\pm$0.50} & \mbox{69.57$\pm$0.97} \\
FedGTA & \mbox{41.72$\pm$1.52} & \mbox{34.69$\pm$2.14} & \mbox{60.88$\pm$4.35} & \mbox{57.90$\pm$5.12} & \mbox{70.33$\pm$1.47} & \mbox{69.29$\pm$1.59} \\
\rowcolor{TableStripe} FedAux-1R & \mbox{34.95$\pm$7.02} & \mbox{21.96$\pm$9.35} & \mbox{52.48$\pm$7.98} & \mbox{49.83$\pm$8.26} & \mbox{62.07$\pm$6.48} & \mbox{59.15$\pm$8.02} \\
\midrule
\rowcolor{PaceHighlight} \textbf{PACE (Ours)} & \mbox{\textbf{74.88$\pm$0.49}} & \mbox{\textbf{74.48$\pm$0.50}} & \mbox{\textbf{68.69$\pm$0.59}} & \mbox{\textbf{66.31$\pm$0.63}} & \mbox{\textbf{83.75$\pm$0.08}} & \mbox{\textbf{83.72$\pm$0.09}} \\
\bottomrule
\end{tabularx}
\vspace{2pt}
\begin{tabularx}{\textwidth}{l*{6}{Y}}
\toprule
\rowcolor{TableHeader}
Methods & \multicolumn{2}{c}{\textbf{CS}} & \multicolumn{2}{c}{\textbf{Computers}} & \multicolumn{2}{c}{\textbf{ogbn-arxiv}} \\
\rowcolor{TableHeader}
 & Accuracy & W-F1 & Accuracy & W-F1 & Accuracy & W-F1 \\
\midrule
Local & \mbox{\underline{85.20$\pm$0.09}} & \mbox{\underline{85.06$\pm$0.09}} & \mbox{\textbf{85.48$\pm$0.20}} & \mbox{\textbf{85.35$\pm$0.23}} & \mbox{\textbf{64.93$\pm$0.13}} & \mbox{\textbf{62.89$\pm$0.14}} \\
\midrule
FedAvg & \mbox{57.82$\pm$2.88} & \mbox{49.12$\pm$3.82} & \mbox{33.87$\pm$14.31} & \mbox{22.30$\pm$12.25} & \mbox{15.84$\pm$1.87} & \mbox{7.78$\pm$2.68} \\
\rowcolor{TableStripe} FedProx & \mbox{64.15$\pm$1.32} & \mbox{57.55$\pm$1.75} & \mbox{38.25$\pm$12.49} & \mbox{27.65$\pm$11.71} & \mbox{23.99$\pm$3.64} & \mbox{18.02$\pm$3.56} \\
FedNova & \mbox{45.53$\pm$7.51} & \mbox{36.77$\pm$7.96} & \mbox{36.87$\pm$0.14} & \mbox{20.63$\pm$0.30} & \mbox{13.15$\pm$3.50} & \mbox{4.17$\pm$1.44} \\
\rowcolor{TableStripe} FedRCL & \mbox{25.34$\pm$5.14} & \mbox{20.34$\pm$2.95} & \mbox{37.68$\pm$2.18} & \mbox{22.18$\pm$3.67} & \mbox{7.12$\pm$5.15} & \mbox{1.39$\pm$1.38} \\
FedPub & \mbox{83.85$\pm$0.61} & \mbox{83.44$\pm$0.68} & \mbox{83.15$\pm$0.78} & \mbox{82.55$\pm$0.93} & \mbox{57.12$\pm$1.04} & \mbox{52.66$\pm$1.42} \\
\rowcolor{TableStripe} FedTAD & \mbox{57.94$\pm$1.86} & \mbox{49.50$\pm$2.63} & \mbox{42.65$\pm$4.29} & \mbox{29.00$\pm$5.76} & \mbox{13.69$\pm$0.47} & \mbox{4.37$\pm$0.85} \\
FedGTA & \mbox{80.72$\pm$0.58} & \mbox{79.89$\pm$0.60} & \mbox{65.12$\pm$5.75} & \mbox{60.71$\pm$4.56} & \mbox{52.53$\pm$0.40} & \mbox{46.98$\pm$0.49} \\
\rowcolor{TableStripe} FedAux-1R & \mbox{57.13$\pm$12.66} & \mbox{51.29$\pm$14.80} & \mbox{58.86$\pm$10.45} & \mbox{53.94$\pm$13.94} & \mbox{50.27$\pm$6.01} & \mbox{46.03$\pm$6.44} \\
\midrule
\rowcolor{PaceHighlight} \textbf{PACE (Ours)} & \mbox{\textbf{86.53$\pm$0.16}} & \mbox{\textbf{86.23$\pm$0.15}} & \mbox{\underline{85.33$\pm$0.12}} & \mbox{\underline{85.10$\pm$0.11}} & \mbox{\textbf{64.93$\pm$0.13}} & \mbox{\textbf{62.89$\pm$0.14}} \\
\bottomrule
\end{tabularx}
\captionof{table}{Complete Louvain results for $C=30$. The layout and notation match the main-paper $C=10$ table.}
\label{tab:full_primary_c30}
\par\medskip

\clearpage
\centering
\scriptsize
\setlength{\tabcolsep}{3pt}
\renewcommand{\arraystretch}{0.80}
\begin{tabularx}{\textwidth}{l*{6}{Y}}
\toprule
\rowcolor{TableHeader}
\multicolumn{7}{c}{\textbf{$C=10$}} \\
\rowcolor{TableHeader}
Methods & Cora & CiteSeer & PubMed & CS & Computers & ogbn-arxiv \\
\midrule
\rowcolor{TableStripe} Local & 78.16$\pm$.25 & 62.01$\pm$.71 & 83.74$\pm$.06 & 86.48$\pm$.15 & 86.16$\pm$.66 & \textbf{42.16$\pm$.28} \\
FedAvg & 13.03$\pm$1.68 & 63.68$\pm$.69 & 72.93$\pm$2.06 & 54.85$\pm$1.82 & 8.90$\pm$1.93 & 7.15$\pm$.98 \\
\rowcolor{TableStripe} FedProx & 23.31$\pm$3.77 & 64.36$\pm$.52 & 61.99$\pm$2.64 & 57.21$\pm$.53 & 15.14$\pm$5.99 & 7.99$\pm$1.00 \\
FedNova & 7.11$\pm$.77 & 29.10$\pm$3.43 & 24.72$\pm$5.06 & 18.28$\pm$5.22 & 5.65$\pm$.17 & 0.94$\pm$.29 \\
\rowcolor{TableStripe} FedRCL & 10.52$\pm$3.67 & 16.44$\pm$3.36 & 17.38$\pm$3.29 & 2.97$\pm$3.31 & 7.49$\pm$2.70 & 0.41$\pm$.26 \\
FedPub & 75.90$\pm$.80 & 65.21$\pm$1.03 & 80.47$\pm$2.18 & 85.83$\pm$.43 & 84.80$\pm$1.61 & 28.35$\pm$.36 \\
\rowcolor{TableStripe} FedTAD & 13.92$\pm$1.24 & 63.32$\pm$.53 & 73.69$\pm$2.28 & 54.35$\pm$1.98 & 10.41$\pm$3.12 & 6.07$\pm$1.17 \\
FedGTA & 28.32$\pm$1.64 & 61.85$\pm$.20 & 56.56$\pm$3.00 & 68.68$\pm$1.00 & 28.80$\pm$5.71 & 10.86$\pm$.24 \\
\rowcolor{TableStripe} FedAux-1R & 55.35$\pm$11.58 & 62.77$\pm$1.99 & 54.66$\pm$12.99 & 63.24$\pm$14.03 & 69.36$\pm$14.98 & 25.04$\pm$3.16 \\
\rowcolor{PaceHighlight} \textbf{PACE (Ours)} & \textbf{79.16$\pm$.61} & \textbf{68.02$\pm$.32} & \textbf{84.24$\pm$.08} & \textbf{86.55$\pm$.24} & \textbf{86.35$\pm$.59} & \textbf{42.16$\pm$.28} \\
\bottomrule
\end{tabularx}
\vspace{3pt}
\begin{tabularx}{\textwidth}{l*{6}{Y}}
\toprule
\rowcolor{TableHeader}
\multicolumn{7}{c}{\textbf{$C=20$}} \\
\rowcolor{TableHeader}
Methods & Cora & CiteSeer & PubMed & CS & Computers & ogbn-arxiv \\
\midrule
\rowcolor{TableStripe} Local & 74.01$\pm$.40 & 57.49$\pm$.25 & 81.71$\pm$.08 & 83.42$\pm$.11 & \textbf{83.93$\pm$1.40} & \textbf{43.01$\pm$.20} \\
FedAvg & 6.66$\pm$.15 & 54.33$\pm$6.39 & 66.18$\pm$2.83 & 41.34$\pm$3.71 & 9.74$\pm$2.05 & 2.36$\pm$.48 \\
\rowcolor{TableStripe} FedProx & 7.34$\pm$.62 & 56.83$\pm$1.81 & 57.06$\pm$3.39 & 48.60$\pm$1.37 & 11.78$\pm$1.86 & 4.72$\pm$1.01 \\
FedNova & 7.64$\pm$.58 & 25.00$\pm$2.43 & 24.46$\pm$4.94 & 19.79$\pm$4.83 & 5.72$\pm$.13 & 0.80$\pm$.21 \\
\rowcolor{TableStripe} FedRCL & 14.26$\pm$2.93 & 17.49$\pm$3.10 & 15.92$\pm$4.03 & 4.64$\pm$2.18 & 7.41$\pm$2.54 & 0.34$\pm$.23 \\
FedPub & 70.30$\pm$1.54 & 57.73$\pm$.57 & 77.48$\pm$1.79 & 79.78$\pm$1.68 & 79.28$\pm$1.64 & 28.91$\pm$2.11 \\
\rowcolor{TableStripe} FedTAD & 6.85$\pm$.33 & 54.06$\pm$6.64 & 68.49$\pm$2.17 & 41.63$\pm$4.77 & 10.09$\pm$1.85 & 1.24$\pm$.52 \\
FedGTA & 40.24$\pm$.72 & 47.67$\pm$6.75 & 69.71$\pm$3.15 & 72.59$\pm$.48 & 36.83$\pm$1.92 & 23.43$\pm$.55 \\
\rowcolor{TableStripe} FedAux-1R & 39.84$\pm$14.83 & 52.73$\pm$3.54 & 60.30$\pm$8.44 & 41.85$\pm$12.52 & 49.90$\pm$20.04 & 26.46$\pm$4.88 \\
\rowcolor{PaceHighlight} \textbf{PACE (Ours)} & \textbf{74.76$\pm$.61} & \textbf{64.35$\pm$.89} & \textbf{82.78$\pm$.13} & \textbf{84.32$\pm$.15} & 83.08$\pm$1.59 & \textbf{43.01$\pm$.20} \\
\bottomrule
\end{tabularx}
\vspace{3pt}
\begin{tabularx}{\textwidth}{l*{6}{Y}}
\toprule
\rowcolor{TableHeader}
\multicolumn{7}{c}{\textbf{$C=30$}} \\
\rowcolor{TableHeader}
Methods & Cora & CiteSeer & PubMed & CS & Computers & ogbn-arxiv \\
\midrule
\rowcolor{TableStripe} Local & 70.15$\pm$.43 & 55.27$\pm$.27 & 81.67$\pm$.13 & 81.03$\pm$.15 & \textbf{84.07$\pm$.53} & \textbf{42.03$\pm$.27} \\
FedAvg & 6.51$\pm$.00 & 54.62$\pm$3.76 & 61.17$\pm$1.83 & 30.55$\pm$2.51 & 8.59$\pm$4.94 & 2.14$\pm$1.02 \\
\rowcolor{TableStripe} FedProx & 6.51$\pm$.00 & 56.24$\pm$1.54 & 52.51$\pm$2.06 & 38.18$\pm$1.43 & 10.44$\pm$5.07 & 4.56$\pm$1.31 \\
FedNova & 8.15$\pm$1.06 & 24.94$\pm$1.99 & 25.22$\pm$4.40 & 19.34$\pm$4.98 & 5.86$\pm$.20 & 0.83$\pm$.25 \\
\rowcolor{TableStripe} FedRCL & 13.85$\pm$2.04 & 20.73$\pm$2.42 & 17.37$\pm$3.28 & 8.82$\pm$1.84 & 6.83$\pm$2.22 & 0.39$\pm$.22 \\
FedPub & 67.39$\pm$1.09 & 55.50$\pm$.63 & 74.85$\pm$3.67 & 77.65$\pm$2.28 & 80.53$\pm$1.45 & 27.46$\pm$1.58 \\
\rowcolor{TableStripe} FedTAD & 6.51$\pm$.00 & 52.99$\pm$5.80 & 63.98$\pm$1.56 & 31.04$\pm$1.84 & 9.88$\pm$2.70 & 0.93$\pm$.18 \\
FedGTA & 27.33$\pm$1.87 & 52.21$\pm$4.91 & 67.62$\pm$1.48 & 72.25$\pm$.46 & 40.50$\pm$2.06 & 22.53$\pm$.28 \\
\rowcolor{TableStripe} FedAux-1R & 15.52$\pm$10.74 & 45.20$\pm$7.61 & 55.76$\pm$8.90 & 35.16$\pm$16.96 & 43.08$\pm$13.23 & 22.54$\pm$5.79 \\
\rowcolor{PaceHighlight} \textbf{PACE (Ours)} & \textbf{72.67$\pm$.46} & \textbf{61.46$\pm$.66} & \textbf{83.16$\pm$.13} & \textbf{81.93$\pm$.18} & 83.49$\pm$.29 & \textbf{42.03$\pm$.27} \\
\bottomrule
\end{tabularx}
\captionof{table}{Complete fixed-class Macro-F1 baseline matrices under Louvain partitioning for $C=10$, $20$, and $30$. Datasets are columns and methods are rows.}
\label{tab:full_macro_f1}

%% file: tab/partition_robustness.tex


\begin{minipage}{\textwidth}
\centering
\scriptsize
\setlength{\tabcolsep}{3pt}
\renewcommand{\arraystretch}{0.80}
\begin{tabularx}{\textwidth}{l*{6}{Y}}
\toprule
\rowcolor{TableHeader}
Methods & \multicolumn{2}{c}{\textbf{Cora}} & \multicolumn{2}{c}{\textbf{CiteSeer}} & \multicolumn{2}{c}{\textbf{PubMed}} \\
\rowcolor{TableHeader}
 & Accuracy & W-F1 & Accuracy & W-F1 & Accuracy & W-F1 \\
\midrule
\rowcolor{TableStripe} Local & \mbox{\underline{78.36$\pm$0.22}} & \mbox{\underline{78.34$\pm$0.22}} & \mbox{68.74$\pm$0.45} & \mbox{\underline{68.33$\pm$0.43}} & \mbox{82.70$\pm$0.11} & \mbox{82.67$\pm$0.11} \\
FedAvg & \mbox{30.80$\pm$0.45} & \mbox{15.63$\pm$0.82} & \mbox{69.83$\pm$0.97} & \mbox{67.21$\pm$0.89} & \mbox{\underline{85.18$\pm$0.26}} & \mbox{\underline{85.07$\pm$0.28}} \\
\rowcolor{TableStripe} FedProx & \mbox{32.83$\pm$0.77} & \mbox{19.40$\pm$1.36} & \mbox{\underline{70.02$\pm$0.24}} & \mbox{67.48$\pm$0.19} & \mbox{84.46$\pm$0.36} & \mbox{84.29$\pm$0.41} \\
FedNova & \mbox{29.87$\pm$0.15} & \mbox{14.08$\pm$0.35} & \mbox{36.77$\pm$1.74} & \mbox{31.27$\pm$2.58} & \mbox{38.97$\pm$2.84} & \mbox{30.85$\pm$4.55} \\
\rowcolor{TableStripe} FedRCL & \mbox{24.29$\pm$6.78} & \mbox{14.82$\pm$3.93} & \mbox{24.16$\pm$2.43} & \mbox{15.66$\pm$2.71} & \mbox{32.02$\pm$10.22} & \mbox{16.29$\pm$8.30} \\
FedPub & \mbox{75.35$\pm$3.56} & \mbox{75.25$\pm$3.76} & \mbox{68.75$\pm$0.69} & \mbox{68.01$\pm$0.79} & \mbox{81.20$\pm$2.35} & \mbox{81.08$\pm$2.49} \\
\rowcolor{TableStripe} FedTAD & \mbox{31.50$\pm$1.00} & \mbox{16.92$\pm$1.78} & \mbox{69.85$\pm$1.04} & \mbox{67.20$\pm$0.97} & \mbox{84.91$\pm$0.26} & \mbox{84.80$\pm$0.28} \\
FedGTA & \mbox{44.53$\pm$0.59} & \mbox{35.58$\pm$0.76} & \mbox{67.74$\pm$1.88} & \mbox{65.11$\pm$1.78} & \mbox{83.76$\pm$0.83} & \mbox{83.67$\pm$0.83} \\
\rowcolor{TableStripe} FedAux-1R & \mbox{60.32$\pm$13.94} & \mbox{57.59$\pm$17.22} & \mbox{67.59$\pm$1.94} & \mbox{65.44$\pm$2.38} & \mbox{79.99$\pm$1.53} & \mbox{79.80$\pm$1.68} \\
\rowcolor{PaceHighlight} \textbf{PACE (Ours)} & \mbox{\textbf{79.68$\pm$0.25}} & \mbox{\textbf{79.49$\pm$0.25}} & \mbox{\textbf{73.46$\pm$0.66}} & \mbox{\textbf{71.83$\pm$0.77}} & \mbox{\textbf{85.31$\pm$0.09}} & \mbox{\textbf{85.30$\pm$0.10}} \\
\bottomrule
\end{tabularx}
\vspace{2pt}
\begin{tabularx}{\textwidth}{l*{4}{Y}}
\toprule
\rowcolor{TableHeader}
Methods & \multicolumn{2}{c}{\textbf{CS}} & \multicolumn{2}{c}{\textbf{Computers}} \\
\rowcolor{TableHeader}
 & Accuracy & W-F1 & Accuracy & W-F1 \\
\midrule
\rowcolor{TableStripe} Local & \mbox{\underline{88.60$\pm$0.14}} & \mbox{\underline{88.56$\pm$0.14}} & \mbox{\underline{86.92$\pm$0.15}} & \mbox{\underline{86.81$\pm$0.13}} \\
FedAvg & \mbox{60.61$\pm$2.17} & \mbox{51.96$\pm$2.91} & \mbox{39.59$\pm$9.82} & \mbox{27.65$\pm$6.78} \\
\rowcolor{TableStripe} FedProx & \mbox{63.97$\pm$2.17} & \mbox{56.57$\pm$3.29} & \mbox{40.32$\pm$11.39} & \mbox{28.82$\pm$9.29} \\
FedNova & \mbox{42.21$\pm$7.73} & \mbox{32.79$\pm$8.33} & \mbox{37.11$\pm$0.08} & \mbox{20.67$\pm$0.32} \\
\rowcolor{TableStripe} FedRCL & \mbox{9.44$\pm$7.40} & \mbox{3.38$\pm$4.06} & \mbox{38.64$\pm$2.75} & \mbox{22.99$\pm$4.22} \\
FedPub & \mbox{88.32$\pm$0.20} & \mbox{88.27$\pm$0.19} & \mbox{86.29$\pm$0.65} & \mbox{85.90$\pm$1.03} \\
\rowcolor{TableStripe} FedTAD & \mbox{61.74$\pm$1.88} & \mbox{53.61$\pm$2.42} & \mbox{40.90$\pm$8.79} & \mbox{29.54$\pm$6.52} \\
FedGTA & \mbox{85.93$\pm$0.68} & \mbox{85.35$\pm$1.03} & \mbox{59.03$\pm$4.48} & \mbox{54.06$\pm$2.51} \\
\rowcolor{TableStripe} FedAux-1R & \mbox{73.12$\pm$3.55} & \mbox{70.01$\pm$4.68} & \mbox{79.91$\pm$6.65} & \mbox{77.87$\pm$8.89} \\
\rowcolor{PaceHighlight} \textbf{PACE (Ours)} & \mbox{\textbf{89.04$\pm$0.12}} & \mbox{\textbf{88.92$\pm$0.13}} & \mbox{\textbf{87.03$\pm$0.07}} & \mbox{\textbf{86.89$\pm$0.11}} \\
\bottomrule
\end{tabularx}
\captionof{table}{Results under METIS partitioning with $C=10$. Values are five-seed mean$\pm$sample standard deviation in percent. Bold and underline mark the best and second-best distinct displayed means.}
\label{tab:metis_results_c10}
\end{minipage}

\begin{minipage}{\textwidth}
\centering
\scriptsize
\setlength{\tabcolsep}{3pt}
\renewcommand{\arraystretch}{0.80}
\begin{tabularx}{\textwidth}{l*{6}{Y}}
\toprule
\rowcolor{TableHeader}
Methods & \multicolumn{2}{c}{\textbf{Cora}} & \multicolumn{2}{c}{\textbf{CiteSeer}} & \multicolumn{2}{c}{\textbf{PubMed}} \\
\rowcolor{TableHeader}
 & Accuracy & W-F1 & Accuracy & W-F1 & Accuracy & W-F1 \\
\midrule
\rowcolor{TableStripe} Local & \mbox{\underline{73.85$\pm$0.33}} & \mbox{\underline{73.32$\pm$0.42}} & \mbox{\underline{79.26$\pm$0.55}} & \mbox{\underline{79.00$\pm$0.58}} & \mbox{\underline{91.71$\pm$0.05}} & \mbox{\underline{91.71$\pm$0.05}} \\
FedAvg & \mbox{34.18$\pm$0.94} & \mbox{21.74$\pm$1.60} & \mbox{40.40$\pm$3.59} & \mbox{36.46$\pm$3.64} & \mbox{81.90$\pm$2.30} & \mbox{81.69$\pm$2.42} \\
\rowcolor{TableStripe} FedProx & \mbox{37.49$\pm$1.91} & \mbox{27.00$\pm$2.64} & \mbox{50.00$\pm$1.08} & \mbox{47.76$\pm$1.65} & \mbox{78.56$\pm$3.37} & \mbox{78.16$\pm$3.61} \\
FedNova & \mbox{29.64$\pm$0.63} & \mbox{16.12$\pm$0.71} & \mbox{32.39$\pm$0.87} & \mbox{27.95$\pm$1.59} & \mbox{38.63$\pm$2.71} & \mbox{29.89$\pm$4.85} \\
\rowcolor{TableStripe} FedRCL & \mbox{21.74$\pm$9.46} & \mbox{12.72$\pm$6.66} & \mbox{23.63$\pm$3.48} & \mbox{15.55$\pm$4.19} & \mbox{35.70$\pm$8.31} & \mbox{19.29$\pm$6.77} \\
FedPub & \mbox{72.95$\pm$1.08} & \mbox{72.03$\pm$1.24} & \mbox{78.56$\pm$0.69} & \mbox{78.17$\pm$0.56} & \mbox{85.96$\pm$3.42} & \mbox{85.91$\pm$3.47} \\
\rowcolor{TableStripe} FedTAD & \mbox{36.64$\pm$1.20} & \mbox{25.67$\pm$1.61} & \mbox{43.07$\pm$4.33} & \mbox{38.62$\pm$4.75} & \mbox{80.67$\pm$3.28} & \mbox{80.35$\pm$3.48} \\
FedGTA & \mbox{52.13$\pm$0.45} & \mbox{46.61$\pm$0.67} & \mbox{63.23$\pm$0.82} & \mbox{56.54$\pm$1.02} & \mbox{62.18$\pm$1.67} & \mbox{53.51$\pm$3.34} \\
\rowcolor{TableStripe} FedAux-1R & \mbox{52.10$\pm$7.50} & \mbox{44.47$\pm$10.94} & \mbox{55.72$\pm$14.48} & \mbox{50.75$\pm$15.64} & \mbox{79.16$\pm$8.13} & \mbox{78.05$\pm$9.50} \\
\rowcolor{PaceHighlight} \textbf{PACE (Ours)} & \mbox{\textbf{77.43$\pm$0.30}} & \mbox{\textbf{76.78$\pm$0.35}} & \mbox{\textbf{81.70$\pm$0.63}} & \mbox{\textbf{81.02$\pm$0.59}} & \mbox{\textbf{92.08$\pm$0.04}} & \mbox{\textbf{92.09$\pm$0.04}} \\
\bottomrule
\end{tabularx}
\vspace{2pt}
\begin{tabularx}{\textwidth}{l*{4}{Y}}
\toprule
\rowcolor{TableHeader}
Methods & \multicolumn{2}{c}{\textbf{CS}} & \multicolumn{2}{c}{\textbf{Computers}} \\
\rowcolor{TableHeader}
 & Accuracy & W-F1 & Accuracy & W-F1 \\
\midrule
\rowcolor{TableStripe} Local & \mbox{\underline{90.32$\pm$0.10}} & \mbox{\underline{90.24$\pm$0.12}} & \mbox{\underline{86.63$\pm$0.16}} & \mbox{\underline{86.54$\pm$0.15}} \\
FedAvg & \mbox{79.91$\pm$0.55} & \mbox{77.13$\pm$0.46} & \mbox{41.77$\pm$3.64} & \mbox{27.46$\pm$5.14} \\
\rowcolor{TableStripe} FedProx & \mbox{82.14$\pm$0.68} & \mbox{79.45$\pm$0.89} & \mbox{43.44$\pm$2.52} & \mbox{30.07$\pm$2.67} \\
FedNova & \mbox{50.43$\pm$3.83} & \mbox{40.05$\pm$3.43} & \mbox{38.01$\pm$0.69} & \mbox{22.79$\pm$1.36} \\
\rowcolor{TableStripe} FedRCL & \mbox{9.39$\pm$6.53} & \mbox{3.91$\pm$5.58} & \mbox{37.33$\pm$0.30} & \mbox{20.82$\pm$1.01} \\
FedPub & \mbox{84.35$\pm$1.77} & \mbox{83.69$\pm$1.90} & \mbox{84.22$\pm$3.33} & \mbox{83.83$\pm$3.69} \\
\rowcolor{TableStripe} FedTAD & \mbox{79.55$\pm$0.80} & \mbox{76.70$\pm$0.77} & \mbox{42.12$\pm$2.60} & \mbox{28.34$\pm$3.35} \\
FedGTA & \mbox{80.37$\pm$0.45} & \mbox{77.76$\pm$0.62} & \mbox{49.03$\pm$6.01} & \mbox{37.68$\pm$7.96} \\
\rowcolor{TableStripe} FedAux-1R & \mbox{82.63$\pm$4.43} & \mbox{81.29$\pm$5.26} & \mbox{80.20$\pm$5.91} & \mbox{79.72$\pm$6.45} \\
\rowcolor{PaceHighlight} \textbf{PACE (Ours)} & \mbox{\textbf{91.48$\pm$0.22}} & \mbox{\textbf{91.37$\pm$0.24}} & \mbox{\textbf{86.66$\pm$0.10}} & \mbox{\textbf{86.55$\pm$0.10}} \\
\bottomrule
\end{tabularx}
\captionof{table}{Results under Dirichlet partitioning with $\alpha=0.5$ and $C=10$. Values are five-seed mean$\pm$sample standard deviation in percent. Bold and underline mark the best and second-best distinct displayed means.}
\label{tab:dirichlet_results_a05_c10}
\end{minipage}

%% file: sec/cnll_theory.tex
\section{Information-Geometric Interpretation of CNLL}
\label{app:cnll_theory}

CNLL is a one-dimensional receiver-local gate rather than a second learned model. The following results characterize the same computation used in the released resolver. Fix a receiver $i$ with nonempty validation set $V_i^{\mathrm{val}}$. For node $v$ and class $c$, write
\begin{equation}
p_{i,v}^{L}(c)=\operatorname{softmax}(z_{i,v}^{L})_c,
\quad
p_{i,v}^{E}(c)=\operatorname{softmax}(z_{i,v}^{E})_c,
\end{equation}
let $d_{i,v}=z_{i,v}^{E}-z_{i,v}^{L}$, and define
$p_{i,v}^{(\alpha)}=\operatorname{softmax}(z_{i,v}^{L}+\alpha d_{i,v})$.

\paragraph{Proposition 1 (information-geometric path).}
For every $\alpha\in[0,1]$, logit interpolation is exactly the normalized geometric opinion pool
\begin{equation}
p_{i,v}^{(\alpha)}(c)=
\frac{
 [p_{i,v}^{L}(c)]^{1-\alpha}
 [p_{i,v}^{E}(c)]^{\alpha}
}{
 \sum_{k=1}^{C}
 [p_{i,v}^{L}(k)]^{1-\alpha}
 [p_{i,v}^{E}(k)]^{\alpha}
}.
\label{eq:supp_geometric_pool}
\end{equation}
Equivalently, it is the weighted reverse-KL barycenter
\begin{align}
p_{i,v}^{(\alpha)}
=\arg\min_{q\in\Delta^{C-1}}\;&
(1-\alpha)D_{\mathrm{KL}}(q\|p_{i,v}^{L}) \notag\\
&+\alpha D_{\mathrm{KL}}(q\|p_{i,v}^{E}).
\label{eq:supp_kl_barycenter}
\end{align}

\paragraph{Proof.}
Substituting the two softmax distributions into the numerator of Eq.~\eqref{eq:supp_geometric_pool} gives
$\exp((1-\alpha)z_{i,v,c}^{L}+\alpha z_{i,v,c}^{E})$
times a class-independent normalizer, which cancels in the denominator. For Eq.~\eqref{eq:supp_kl_barycenter}, expand both KL terms and impose $\sum_c q(c)=1$ with a Lagrange multiplier. Stationarity gives
$\log q(c)=\mathrm{const}+(1-\alpha)\log p_{i,v}^{L}(c)+\alpha\log p_{i,v}^{E}(c)$;
normalization yields Eq.~\eqref{eq:supp_geometric_pool}.

\paragraph{Corollary (logit-gauge invariance).}
Adding any class-independent constants to $z_{i,v}^{L}$ or $z_{i,v}^{E}$ leaves the entire path $p_{i,v}^{(\alpha)}$ unchanged. Such shifts only multiply the numerator and denominator of Eq.~\eqref{eq:supp_geometric_pool} by the same factor. Thus the gate depends on predictive beliefs, not on an arbitrary softmax logit origin.

\paragraph{Proposition 2 (convexity and moment matching).}
The receiver's validation risk is
\begin{equation}
\mathcal{R}_i(\alpha)=
-\frac{1}{|V_i^{\mathrm{val}}|}
\sum_{v\in V_i^{\mathrm{val}}}
\log p_{i,v}^{(\alpha)}(y_v).
\label{eq:supp_cnll_risk}
\end{equation}
Its first two derivatives are
\begin{align}
\mathcal{R}_i'(\alpha)
&=\frac{1}{|V_i^{\mathrm{val}}|}
\sum_{v\in V_i^{\mathrm{val}}}
\left(
\mathbb{E}_{c\sim p_{i,v}^{(\alpha)}}[d_{i,v,c}]
-d_{i,v,y_v}
\right), \label{eq:supp_cnll_first}\\
\mathcal{R}_i''(\alpha)
&=\frac{1}{|V_i^{\mathrm{val}}|}
\sum_{v\in V_i^{\mathrm{val}}}
\operatorname{Var}_{c\sim p_{i,v}^{(\alpha)}}[d_{i,v,c}]
\ge 0. \label{eq:supp_cnll_second}
\end{align}
Hence $\mathcal{R}_i$ is globally convex. Every interior minimizer satisfies the moment-matching condition
\begin{equation}
\frac{1}{|V_i^{\mathrm{val}}|}
\sum_{v\in V_i^{\mathrm{val}}}
\mathbb{E}_{c\sim p_{i,v}^{(\alpha_i^\star)}}[d_{i,v,c}]
=
\frac{1}{|V_i^{\mathrm{val}}|}
\sum_{v\in V_i^{\mathrm{val}}} d_{i,v,y_v}.
\label{eq:supp_moment_match}
\end{equation}

\paragraph{Proof.}
Differentiating the log-sum-exp form of Eq.~\eqref{eq:supp_cnll_risk} gives Eq.~\eqref{eq:supp_cnll_first}; differentiating its softmax expectation gives the variance in Eq.~\eqref{eq:supp_cnll_second}. Convexity justifies the released endpoint tests: $\mathcal{R}_i'(0)\ge0$ selects Local, $\mathcal{R}_i'(1)\le0$ selects External, and otherwise bisection locates an interior root. Strict convexity is not required; in a degenerate flat segment the minimizer need not be unique.

\paragraph{Implementation correspondence.}
Equation~\eqref{eq:supp_cnll_first} is exactly the derivative evaluated by the released resolver: its first term is the model-implied collaborative displacement and its second is the label-observed displacement. The two endpoint evaluations either certify a boundary optimum or bracket a zero of this monotone derivative. Sixty-four bisection iterations then solve the single scalar problem; they do not update model parameters or consult test labels.

\paragraph{Proposition 3 (Validation-NLL no-regret).}
Define the cumulative validation log-evidence relative to Local as
\begin{equation}
\mathcal{E}_i(\alpha)=
\sum_{v\in V_i^{\mathrm{val}}}
\log
\frac{p_{i,v}^{(\alpha)}(y_v)}
     {p_{i,v}^{L}(y_v)}.
\end{equation}
Then maximizing $\mathcal{E}_i$ is equivalent to minimizing $\mathcal{R}_i$, because
\begin{equation}
\mathcal{E}_i(\alpha)=
-|V_i^{\mathrm{val}}|
\left[\mathcal{R}_i(\alpha)-\mathcal{R}_i(0)\right].
\end{equation}
Since $\alpha=0$ is feasible,
\begin{equation}
\mathcal{E}_i(\alpha_i^\star)\ge0,
\qquad
\mathcal{R}_i(\alpha_i^\star)\le\mathcal{R}_i(0).
\label{eq:supp_validation_no_regret}
\end{equation}
This is a receiver-local, aggregate validation-NLL guarantee. It does not imply a test-accuracy guarantee, a per-node improvement, or a worst-client guarantee. If $V_i^{\mathrm{val}}$ is empty, the protocol defines $\alpha_i^\star=0$ directly.